\documentclass[10pt,twocolumn,letterpaper]{article}

\usepackage{cvpr}              
\usepackage{colortbl}
\usepackage{xcolor}
\usepackage{bm}      
\usepackage{multirow}      
\usepackage{microtype}
\usepackage{cuted}

\definecolor{cvprblue}{rgb}{0.21,0.49,0.74}
\usepackage[pagebackref,breaklinks,colorlinks,allcolors=cvprblue]{hyperref}

\def\paperID{} 
\def\confName{CVPR}
\def\confYear{2025}

\title{Free360:  Layered Gaussian Splatting for Unbounded 360-Degree View Synthesis from  Extremely Sparse and Unposed Views}

\author{
Chong Bao$^{1}$\footnotemark[4]\quad
Xiyu Zhang$^{1}$ \quad
Zehao Yu$^{3}$  \quad
Jiale Shi$^{1}$ \quad
Guofeng Zhang$^{1}$ \quad \\
Songyou Peng$^{2}$ \quad
Zhaopeng Cui$^{1}$\footnotemark[2] \\
$^{1}$State Key Lab of CAD\&CG, Zhejiang University \quad
$^{2}$ETH Zürich \quad \\ $^{3}$University of Tübingen, Tübingen AI Center\\
}

\def\ie{i.e\onedot}

\definecolor{tabfirst}{rgb}{1, 0.7, 0.7}
\definecolor{tabsecond}{rgb}{1, 0.85, 0.7}
\definecolor{tabthird}{rgb}{1, 1, 0.7}
\newcommand{\ssecspace}{\vspace{-0.5em}}

\begin{document}
\twocolumn[{%
\renewcommand\twocolumn[1][]{#1}%
\maketitle
\vspace{-3.5em}
\begin{center}
    \centering
    \includegraphics[width=1.0\linewidth, trim={0 0 0 0}, clip]{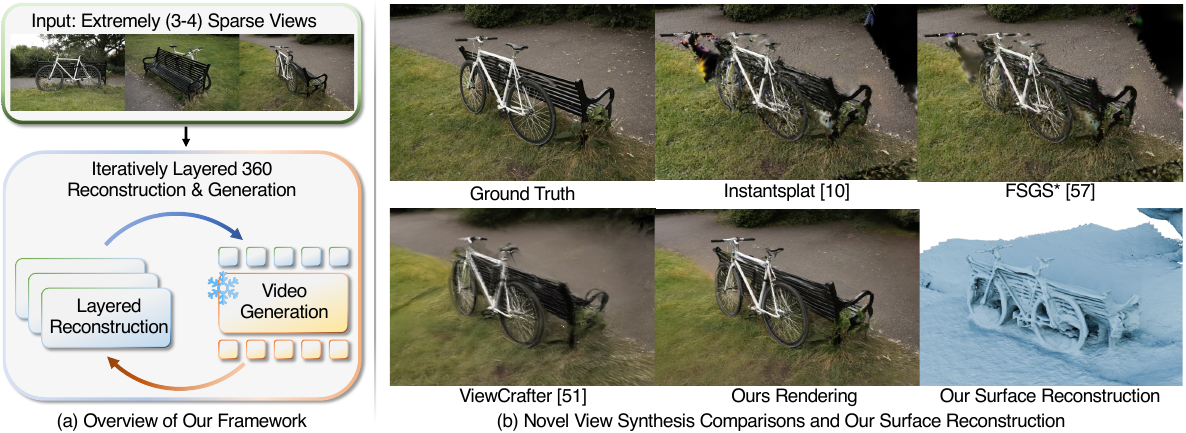}
        \vspace{-2.0em}
    \captionof{figure}{
    \textbf{Free360.}
    (a) We propose a novel Gaussian-based framework, which can reconstruct unbounded 360$^{\circ}$ scenes from extremely (3-4) sparse views through an iterative fusion of layered reconstruction and generation. (b) Our method outperforms other state-of-the-art methods in rendering quality and supports complete surface reconstruction.
    }
    \label{fig:teaser}
\end{center}%
}]

\renewcommand{\thefootnote}{\fnsymbol{footnote}}
\footnotetext[2]{Corresponding author.}
\footnotetext[4]{The work was partially done when visiting ETHZ.}

\maketitle
\begin{abstract}
Neural rendering
has demonstrated remarkable success in high-quality 3D neural reconstruction and novel view synthesis with dense input views and accurate poses.
However, applying it to extremely sparse, unposed views in unbounded 360$^{\circ}$ scenes remains a challenging problem. 
In this paper, we propose a novel neural rendering framework to accomplish the unposed and extremely sparse-view 3D reconstruction in unbounded 360$^{\circ}$ scenes.
To resolve the spatial ambiguity inherent in unbounded scenes with sparse input views, we propose a layered Gaussian-based representation to effectively model the scene with distinct spatial layers.
By employing a dense stereo reconstruction model to recover coarse geometry, we introduce a layer-specific bootstrap optimization to refine the noise and fill occluded regions in the reconstruction.
Furthermore, we propose an iterative fusion of reconstruction and generation alongside an uncertainty-aware training approach to facilitate mutual conditioning and enhancement between these two processes.
Comprehensive experiments show that our approach outperforms existing state-of-the-art methods in terms of rendering quality and surface reconstruction accuracy.
Project page: \href{https://zju3dv.github.io/free360/}{https://zju3dv.github.io/free360/}.

\end{abstract}

\ssecspace
\vspace{-1.0em}

\section{Introduction}
\label{sec:intro}

3D Gaussian Splatting (3DGS)~\cite{3dgs} has shown great success in achieving high efficiency and quality in 3D neural reconstruction and novel view synthesis with dense input views and accurate poses.
With the development of large reconstruction models~\cite{hong2024lrmlargereconstructionmodel, zhang2025gs,xu2024grm} and generative models~\cite{liu2023zero, sargent2024zeronvs,liu2023one2345,yu2024viewcrafter}, it even becomes feasible to recover 3D objects or bounded scenes from sparse 
views.

Most sparse-view 3DGS approaches~\cite{fsgs, dngaussian, loopsparsegs} rely on precise camera poses and relatively sparse sets of views (typically more than 10).  To handle pose-free scenarios, some methods ~\cite{fan2024instantsplat} tend to learn a 3DGS based on the point cloud and camera poses obtained from dense stereo reconstruction models~\cite{dust3r,mast3r}; however, these stereo models often yield noisy geometry in unbounded scenes, leading to more artifacts and corrupted structures at novel viewpoints. 
Other methods~\cite{wang2024motionctrl, feng2025explorative, you2024nvs, yu2024viewcrafter} tempt to fine-tune video diffusion models for direct novel view generation, while such methods are limited to narrow baseline or highly-overlapped scenarios due to the inherent absence of 3D information in the video generation model.
In the unbounded 360$^{\circ}$ scenes with low-overlap sparse views, these generated views exhibit severe multi-view inconsistencies, further contaminating the reconstruction. Therefore, performing neural reconstruction and unbounded 360$^{\circ}$ view synthesis from extremely sparse and unposed views remains an underexplored and challenging problem.

In this paper, we present a novel neural rendering framework to realize unbounded 360$^{\circ}$ view synthesis and 3D reconstruction from extremely sparse views (e.g., 3-4 views) by analyzing and addressing two key challenges in unposed and unbounded scenarios. 
At first, the dense stereo reconstruction model~\cite{dust3r, mast3r} is employed to recover the coarse geometry and camera poses.
However, the large depth span of the unbounded scene and insufficient multi-view correspondences from 360$^{\circ}$ extremely sparse views lead to two critical issues:
1) unreliable depth estimation, which hinders the clear differentiation between near-camera and distant structures, and 2) different visibility characteristics, which pose challenges for unified optimization, as near-camera content appears in multiple views, whereas distant structures are often partially occluded and visible from only a single viewpoint.
To tackle this issue, we propose a layered Gaussian-based representation enabling layer-specific bootstrap optimizations.
Our approach explicitly constructs the scene's layered structure and use a photometric-guided optimization to mitigate noise and correct detail distortions for near-camera reconstruction, alongside a prior-guided inpainting to complete missing regions in distant reconstruction.

Secondly, given the extremely sparse views, intensive observations are missing for 360$^{\circ}$ reconstruction, requiring the use of generative models to generate additional observations.
Furthermore, the latest video diffusion models~\cite{yu2024viewcrafter} have demonstrated the capability of generating novel views from a good condition, e.g., accurate per-frame point cloud renderings, which however are hard to obtain given extremely sparse views.
To tackle this challenge, we propose an iterative fusion strategy that seamlessly integrates reconstruction and generation. 
The generation provides supplementary observations for reconstruction to alleviate the spare-view ambiguity.
In turn, the reconstruction resolves inconsistency within generated views and produces consistent renderings to condition the generation process.
Moreover, to prevent error propagation during the fusion process and ensure a robust reconstruction upon generation, we develop an uncertainty-aware training that filters out inconsistent generated contents.

In summary, the contributions of our paper are as follows. (1) We propose a novel neural rendering framework to accomplish the unposed and extremely sparse-view 3D reconstruction in unbounded 360$^{\circ}$ scenes.
(2) We propose a layered Gaussian-based representation to address the spatial ambiguity of unbounded scenes, along with a layer-specific bootstrap optimization upon the layered representation to refine the noisy reconstruction. 
(3) We design an iterative fusion strategy of reconstruction and novel view generation, facilitating mutual conditioning and
enhancement between the two processes. Besides, we incorporate uncertainty-aware learning to mitigate error propagation and ensure robust reconstruction.
(4) We conduct extensive experiments on various large-scale unbounded 360$^{\circ}$ scenes using only 3-9 views. The results demonstrate that our method reaches the best rendering quality and surface reconstruction accuracy.

\ssecspace

\begin{figure*}[!t]
\centering
\includegraphics[width=0.97\linewidth, trim={0 0 0 0}, clip]{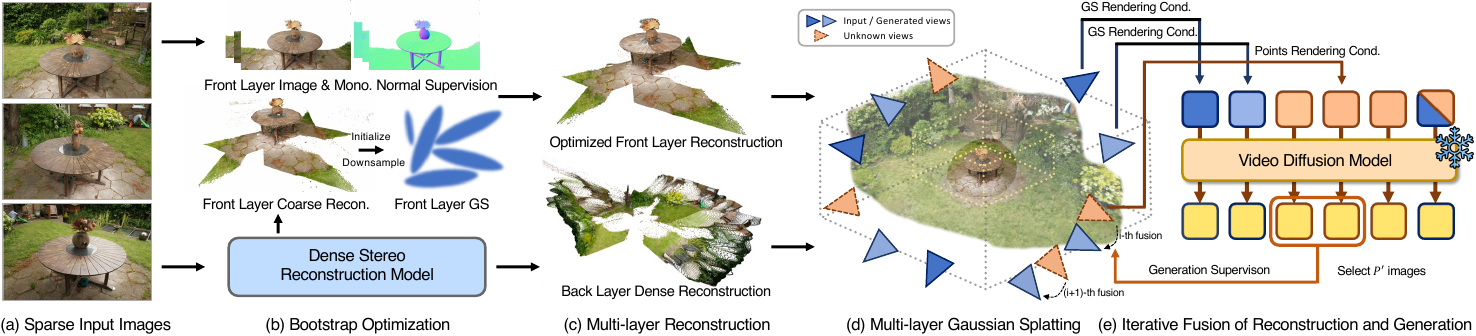}
\vspace{-0.5em}
\caption{
\textbf{Pipeline.}
(a-d) Given unposed extremely sparse views, we employ the dense stereo reconstruction model~\cite{dust3r, mast3r} to recover camera poses and initial point cloud of the scene.
A layered Gaussian-based representation is built upon the initial point cloud to enable layer-specific bootstrap optimization.
(e) We design the iterative fusion of reconstruction and generation with diffusion model~\cite{yu2024viewcrafter}. Unknown views are iteratively generated under conditions of consistent GS rendering of known views. In turn, generated views are used to enhance the GS training.
}
\vspace{-1.0em}
\label{fig:framework}
\end{figure*}

\section{Related Works}
\label{sec:related_works}
\noindent\textbf{Sparse-view Novel View Synthesis.}
Novel view synthesis (NVS) aims to generate new images from viewpoints outside the input views. Since the introduction of NeRF~\cite{nerf}, the NVS field has evolved rapidly, with 3D Gaussian Splatting~\cite{3dgs,Yu2023MipSplatting} emerging as the new standard. While both NeRF and 3DGS produce impressive photometric NVS results from dense input views, their performance degrades significantly with sparse input views due to limited optimization constraints~\cite{barron2022mipnerf360,kaizhang2020}. Subsequent work has aimed to enhance sparse-view NVS by incorporating semantic regularization~\cite{dietnerf, bao2023sine, bao2024geneavatar}, smoothness priors~\cite{regnerf,yang2023freenerf}, or geometric priors~\cite{Yu2022MonoSDF,dsnerf, ddpnerf,fsgs,dngaussian, yang2022neumesh}. For instance, FSGS~\cite{fsgs} and DNGaussian~\cite{dngaussian} improve sparse-view NVS by regularizing depth maps rendered from 3D Gaussians with monocular depth priors~\cite{Ranftl2020,Ranftl2021}. However, these methods either focus on forward-facing scenes~\cite{mildenhall2019llff,aanaes2016large} or rely on relatively dense input views (e.g., 24 views in the Mip-NeRF 360 dataset~\cite{barron2022mipnerf360}). In contrast, we solve an extremely challenging scenario involving only 3 or 4 views for 360$^{\circ}$ inward-facing scenes without assuming ground-truth poses.

\noindent\textbf{Pose-free Novel View Synthesis.}
Optimizing a NeRF or 3DGS model typically requires known camera poses beforehand. In practice, these poses are often recovered using established structure-from-motion (SfM) methods, such as COLMAP~\cite{schoenberger2016sfm}. However, traditional SfM techniques face significant challenges in accurately estimating poses when provided with sparse input views, making it impractical to assume known poses in such scenarios. While pose-free methods~\cite{lin2021barf,cheng2023lu,bian2023nope,cf3dgs} have emerged to tackle this issue by jointly optimizing camera poses and scene representation, these approaches often struggle under sparse input conditions. Recently, DUSt3R~\cite{dust3r}, an unstructured dense stereo reconstruction model, has demonstrated robust performance in estimating dense 3D points and camera poses from sparse views. Building on DUSt3R’s output, InstantSplat~\cite{fan2024instantsplat} performs joint optimization on poses and Gaussian attributes, achieving improved NVS renderings. Following this line of research, we leverage DUSt3R to estimate camera poses and dense 3D points from sparse input views.
However, we observe needle-like artifacts when using the noisy points from DUSt3R.
To address this, we build a layered Gaussian Splatting with layer-specific bootstrap optimization.

\noindent\textbf{Generative Sparse Novel View Synthesis.}
Generative models have achieved remarkable progress in recent years, delivering impressive results in image~\cite{rombach2021highresolution}, video~\cite{videoworldsimulators2024}, and 3D generation~\cite{poole2022dreamfusiontextto3dusing2d}. This success has led to the exploration of generative models for sparse novel view synthesis (NVS)~\cite{xiong2023sparsegs,liu2023zero,sargent2024zeronvs,yang2024gaussianobject,liu2023deceptive,wu2024reconfusion,gao2024cat3d}.
For instance, ReconFusion~\cite{wu2023reconfusion} trains a diffusion model to refine the noisy rendering of a NeRF. With the diffusion model, ReconFusion optimizes a NeRF with a sampling loss from the denoising model for novel views and a reconstruction loss between sparse input views. 
More recent approaches~\cite{wang2024motionctrl,you2024nvs, feng2025explorative, yu2024viewcrafter} fine-tune video diffusion models for novel view synthesis from sparse inputs. Although these methods perform well with input images with significant overlap, they often face identity shift and multi-view inconsistency issues when provided with low-overlapped sparse views.
In our work, we build upon video diffusion models by introducing a progressive fusion step to enhance reconstruction by selecting reliable images generated by the video diffusion model. Furthermore, we compute uncertainty maps to capture inconsistencies in the generated images, guiding the reconstruction process.

\ssecspace

\section{Method}
As shown in Fig.~\ref{fig:framework}, given unposed extremely sparse views in an unbounded 360$^{\circ}$ scene, we propose a novel neural rendering framework to faithfully reconstruct the structure of the scene and render high-quality novel views.
By utilizing the dense stereo reconstruction model~\cite{dust3r,mast3r} to obtain the camera poses and an initial point cloud from sparse views, we partition the scene into a layered structure and build a layered Gaussian-based representation upon it (see Sec.~\ref{subsec:multi_3dgs}).
Then, we introduce a layer-specific bootstrap optimization technique to refine reconstruction error (see Sec.~\ref{subsec:bootstrap}).
Due to the inadequate constraints provided by sparse views, we incorporate the video diffusion model~\cite{yu2024viewcrafter} as prior and propose an iterative fusion approach that combines reconstruction and generation, enabling mutual conditioning and enhancement between the two processes, \ie,  generation provides novel observations for reconstruction while reconstruction resolves inconsistencies within the generated outputs (see Sec.~\ref{subsec:iterative}).
Furthermore, we propose an uncertainty-aware training technique to achieve robust reconstruction with the awareness of inconsistent generated content (see Sec.~\ref{subsec:uncertainty}).

\subsection{Layered Gaussian Splatting}
\label{subsec:multi_3dgs}
In the spirit of partitioning the near-camera and far-away scene content into different layers for layer-specific optimization, existing unbounded NeRF methods~\cite{zhang2020nerf++, barron2022mipnerf360} intuitively treat the scene inside the smallest sphere encompassing all camera positions as foreground, with all other areas as background.
Following this spirit, we use a more compact bounding volume to partition an unbounded scene into a front layer and a back layer.
Specifically, the front layer is the scene inside the smallest bounding sphere that encapsulates the intersection area of all camera frustums, while the back layer constitutes the residual scene.
Alternatively, a more precise partition is to annotate the front and back layers on the monocular depth~\cite{depthanythingv2} of each sparse-input view.

Given unposed sparse images, we use the dense stereo reconstruction model~\cite{dust3r, mast3r} to recover the scene's camera poses and initial point cloud.
Then, the point cloud is partitioned into the front-layer and back-layer point clouds as described above. 
We build our layered-based representation upon the 2D Gaussian Splatting~\cite{huang20242d}.
Each layer $i$ is independently modeled as a group of Gaussian primitives $G_i = \{\mathcal{G}_{i,k} | k=1,... ,K\}$, which is initialized using partitioned point clouds. 
The Gaussian primitives are parameterized by the center position $\mu_{i,k}$,  a scaling vector $S_{i,k}$, and a rotation matrix $R_{i,k}$.
The color $c_{i,k}$ of Gaussian primitives is characterized by the view-dependent spherical harmonics (SH).
To render the entire scene, we merge all Gaussian primitives within each layer and execute Gaussian rasterization once for efficiency and anti-aliasing.
The pixel color is composited by the point-based $\alpha$-blending of rasterized Gaussian that overlaps the pixel.

\subsection{Bootstrap Optimization}
\label{subsec:bootstrap}

\begin{figure}[!t]
\centering
\includegraphics[width=0.97\linewidth, trim={0 0 0 0}, clip]{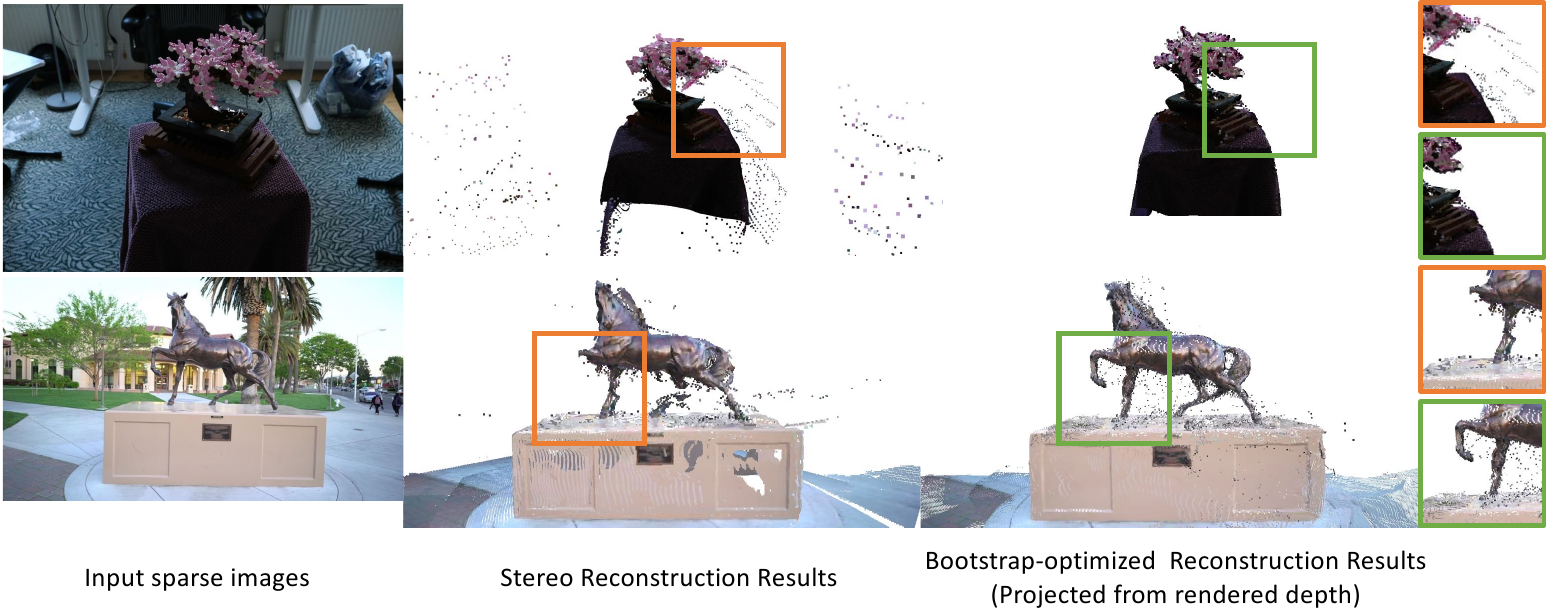}
\vspace{-1.0em}
\caption{\textbf{Visualization of Point Clouds.} We compare results from the stereo reconstruction model and our bootstrap optimization. 
    }
\label{fig:method_bootstrap}
\end{figure}

\begin{figure}[!t]
\centering
\vspace{-1.0em}
\includegraphics[width=0.97\linewidth, trim={0 0 0 0}, clip]{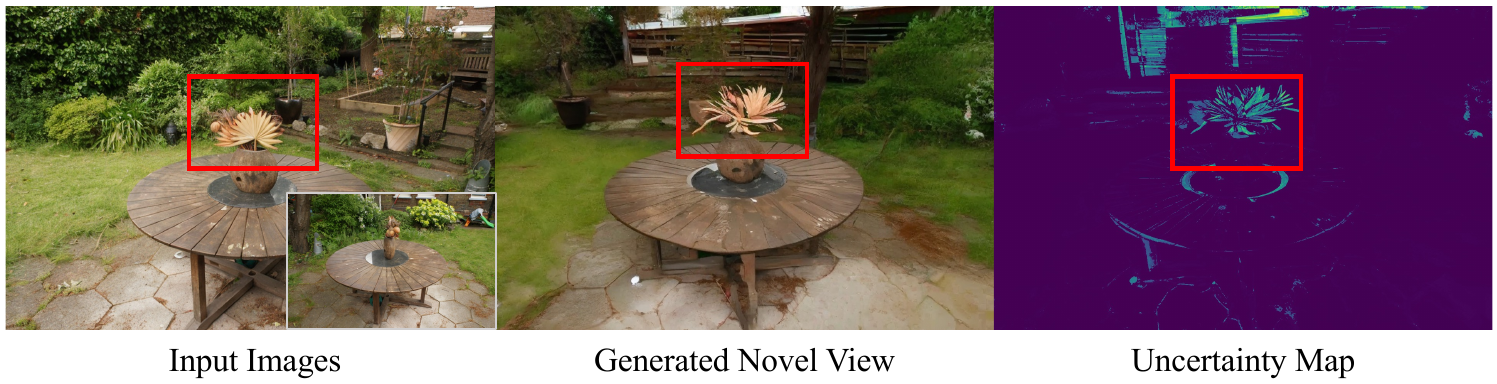}
\vspace{-1.0em}
\caption{\textbf{Uncertainty Map.} We show the uncertainty map estimated from generated novel views. The flower has severe multi-view inconsistency with high uncertainty.
    }
\vspace{-1.4em}
\label{fig:method_uncertainty}
\end{figure}

The point cloud produced by reconstructing an entire scene using a stereo model is always error-prone due to the insufficient visual constraints from sparse views and the model’s inability to perceive the fine structures and occlusion.
With the layered GS, we target layer-specific bootstrap optimization to refine these errors.

For the front layer, a noisy point cloud and missing geometry of the fine structure always occur, such as the noisy floating points of Bonsai and missing leg of Horse in Fig.~\ref{fig:method_bootstrap}.
This can further diminish the quality of Gaussian rendering and generation consistency.
Inspired by RAIN-GS~\cite{jung2024relaxing}, which demonstrates that photometric-based GS optimization can recover clean geometry with fine details when initialized with as few as 10 points, we perform an initial GS optimization on the front layer using a downsampled point cloud.
This downsampling operation avoids initializing considerable error points of the point cloud as Gaussians by employing a voxel-based uniform downsampling strategy.
The range of voxel size is 1\% to 3\% of the bounding box of the point cloud.
Given the sparsity of input views, we further integrate monocular normal constraints~\cite{ye2024stablenormal} to enhance geometric accuracy.
During optimization, we enable densification to prune noisy Gaussians while expanding new Gaussians to capture missing fine structures, guided by the images and monocular normals.
Specifically, we use photometric loss $\mathcal{L}_c$ between the rendered images and input images to capture details of geometry and binary entropy loss $\mathcal{L}_m$ between the rendered alpha map and layer mask to remove the dislocated outlier,
a cosine similarity loss $\mathcal{L}_{n}$ between rendered normal map and monocular normal to fix the bumpy surface and use L1 loss $\mathcal{L}_d$ between the rendered and stereo-reconstructed~\cite{dust3r} depths to regularize the geometry $\mathcal{L}_d$:
\begin{equation}
      \mathcal{L}_{bs} = \mathcal{L}_c + \lambda_m \mathcal{L}_m  + \lambda_n \mathcal{L}_n + \lambda_d \mathcal{L}_d,
\label{eq:bs_loss}
\end{equation}
where $\lambda_m=1.0$, $\lambda_n=0.25$, $\lambda_d=0.1$.
As shown in Fig.~\ref{fig:method_bootstrap}, our optimization successfully gets a cleaner point cloud in Bonsai and fills the missing geometry of leg in Horse.

For the back layer, we inpaint the depth of regions in the back layer that are not visible from the input viewpoints.
To achieve this, we sample several novel viewpoints and render their visibility masks based on the back layer's point cloud.
Then, we obtain the bounding sphere of the back layer's point cloud,
and a ray-box intersection is performed to estimate the depth of invisible pixels.
Based on the inpainted depth, new points are subsequently added to back layer's point cloud.

\begin{table*}[h!]
\vspace{-1.0em}
\centering
\resizebox{0.95\linewidth}{!}{
\tabcolsep 10pt
\begin{tabular}{lccc ccc ccc}
\hline
Method & \multicolumn{3}{c}{3 views} & \multicolumn{3}{c}{6 views} & \multicolumn{3}{c}{9 views} \\
\hline
 & PSNR$\uparrow$ & SSIM$\uparrow$ & LPIPS$\downarrow$ & PSNR$\uparrow$ & SSIM$\uparrow$ & LPIPS$\downarrow$ & PSNR$\uparrow$ & SSIM$\uparrow$ & LPIPS$\downarrow$ \\
\hline
FSGS*~\cite{fsgs} & 14.25 & 0.286 & 0.545 & 15.65 & 0.318 & \cellcolor{tabthird}0.509 & 16.37 & 0.347 & \cellcolor{tabthird}0.489 \\
InstantSplat~\cite{fan2024instantsplat} & 14.10 & 0.296 & \cellcolor{tabthird}0.529 & \cellcolor{tabthird}15.83 & \cellcolor{tabthird}0.351 & \cellcolor{tabsecond}0.480 & \cellcolor{tabsecond}17.16 & \cellcolor{tabsecond}0.402 & \cellcolor{tabsecond}0.443 \\
ZeroNVS*~\cite{sargent2024zeronvs} & \cellcolor{tabthird}15.16 & \cellcolor{tabthird}0.327 & 0.632 & 15.46 & 0.313 & 0.614 & 15.81 & 0.328 & 0.607 \\
ViewCrafter~\cite{yu2024viewcrafter} & \cellcolor{tabsecond}15.63 & \cellcolor{tabsecond}0.358 & \cellcolor{tabsecond}0.515 & \cellcolor{tabsecond}16.32 & \cellcolor{tabsecond}0.366 & 0.555 & \cellcolor{tabthird}16.68 & \cellcolor{tabthird}0.382 & 0.551 \\
\textbf{Ours} & \cellcolor{tabfirst}16.20 & \cellcolor{tabfirst}0.378 & \cellcolor{tabfirst}0.499 & \cellcolor{tabfirst}16.83 & \cellcolor{tabfirst}0.397 & \cellcolor{tabfirst}0.458 & \cellcolor{tabfirst}17.35 & \cellcolor{tabfirst}0.423 & \cellcolor{tabfirst}0.435 \\
\hline
\end{tabular}
}
\vspace{-0.5em}
\caption{\textbf{Comparison on Mip-NeRF 360$^{\circ}$~\cite{barron2022mipnerf360} Dataset.} We compare the rendering quality with baselines given 3, 6 and 9 views.}
\label{tab:comp_mipnerf}
\end{table*}
\begin{table*}[h!]
\vspace{-0.5em}
\centering
\resizebox{0.95\linewidth}{!}{
\tabcolsep 10pt
\begin{tabular}{lccc ccc ccc}
\hline
Method & \multicolumn{3}{c}{4 views} & \multicolumn{3}{c}{6 views} & \multicolumn{3}{c}{9 views} \\
\hline
 & PSNR$\uparrow$ & SSIM$\uparrow$ & LPIPS$\downarrow$ & PSNR$\uparrow$ & SSIM$\uparrow$ & LPIPS$\downarrow$ & PSNR$\uparrow$ & SSIM$\uparrow$ & LPIPS$\downarrow$ \\
\hline
FSGS*~\cite{fsgs} & \cellcolor{tabthird}13.14 & 0.357 & 0.509 & \cellcolor{tabsecond}14.29 & \cellcolor{tabthird}0.425 & \cellcolor{tabthird}0.451 & \cellcolor{tabthird}15.13 & \cellcolor{tabthird}0.457 & \cellcolor{tabthird}0.423 \\
InstantSplat~\cite{fan2024instantsplat} & 13.03 & \cellcolor{tabthird}0.390 & \cellcolor{tabsecond}0.496 & \cellcolor{tabthird}14.22 & \cellcolor{tabsecond}0.457 & \cellcolor{tabsecond}0.439 & \cellcolor{tabsecond}15.58 & \cellcolor{tabsecond}0.51 & \cellcolor{tabsecond}0.391 \\
ZeroNVS*~\cite{sargent2024zeronvs} & 12.62 & 0.363 & 0.619 & 12.96 & 0.349 & 0.597 & 13.18 & 0.364 & 0.592 \\
ViewCrafter~\cite{yu2024viewcrafter} & \cellcolor{tabsecond}13.56 & \cellcolor{tabsecond}0.412 & \cellcolor{tabthird}0.504 & 14.12 & 0.424 & 0.483 & 14.70 & 0.443 & 0.467 \\
\textbf{Ours} & \cellcolor{tabfirst}14.69 & \cellcolor{tabfirst}0.476 & \cellcolor{tabfirst}0.409 & \cellcolor{tabfirst}15.67 & \cellcolor{tabfirst}0.523 & \cellcolor{tabfirst}0.368 & \cellcolor{tabfirst}16.73 & \cellcolor{tabfirst}0.564 & \cellcolor{tabfirst}0.328 \\
\hline
\end{tabular}
}
\vspace{-0.5em}
\caption{\textbf{Comparison on Tanks and Temples~\cite{tanksandtemples} Dataset.} We compare the rendering quality with baselines given 4, 6 and 9 views.}
\vspace{-1.0em}
\label{tab:comp_tnt}
\end{table*}
\subsection{Iterative Fusion of Reconstruction and Generation}
\label{subsec:iterative}
Due to underdetermined constraints in the extremely sparse view, we leverage video diffusion model~\cite{yu2024viewcrafter} to generate new observations for 360$^{\circ}$ reconstruction.
Here we use ViewCrafter~\cite{yu2024viewcrafter} as generative prior.
ViewCrafter~\cite{yu2024viewcrafter} takes $P$ images of a sequential motion as conditions, which consists of point cloud renderings and real images, and generates $P$ novel views following the conditioned motions.
A na\"ive way to use generative prior is to generate all required novel views at once and train Gaussian primitives directly on them.
However, ViewCrafter~\cite{yu2024viewcrafter} exhibits pronounced multi-view inconsistency when generating novel views between low-overlapped sparse views, as it is trained on videos characterized by minimal camera motion and substantial per-frame overlap.
These inconsistencies degrade the GS optimization process, leading to blurry and distorted renderings.

To mitigate the inconsistency in the generation, we propose an iterative fusion strategy of reconstruction and generation where the diffusion model is iteratively conditioned on rendered images and generates consistent novel views to enhance the reconstruction in turn.
Specifically, a known image set $\mathbf{I}_{\text{known}}$ and pose set $\mathbf{V}_{\text{known}}$ are defined to denote the cameras with known image supervision (real or generated).
We initialize the known image set and pose set with input sparse views, $\mathbf{I}_{\text{known}} = \mathbf{I}_{\text{input}}$, $\mathbf{V}_{\text{known}} = \mathbf{V}_{\text{input}}$.
Starting from sampling a start pose among input poses, we interpolate $P$ poses  sequentially $\mathbf{V}_{\text{gen}}$
that consist of known cameras $\mathbf{V}_{\text{known}, \text{gen}}$ (generated at the iteration before) and unknown cameras $\mathbf{V}_{\text{novel}}$, 
$\mathbf{V}_{\text{gen}}=\mathbf{V}_{\text{known}, \text{gen}} \cup \mathbf{V}_{\text{novel}}$.
For the known cameras, we use real images or GS-rendered images as generative conditions.
For unknown cameras, we use
renderings of whole-scene point cloud using
Pytorch3D~\cite{ravi2020pytorch3d} as the generative conditions.
Next, we utilize the video diffusion model to generate images $\mathbf{I}_{\text{gen}}$ based on these conditions,
$\mathbf{I}_{\text{gen}}=\mathbf{I}_{\text{known}, \text{gen}} \cup \mathbf{I}_{\text{novel}}$
.
The video diffusion model~\cite{yu2024viewcrafter} generates $P$ sequential frames at a single forward, but the frame quality is not constant.
The inconsistency is accumulated with distance,
\ie the consistency of novel views
deteriorates as the camera deviates from the known cameras.
Therefore, we define a reliable frame selection strategy.
Only $P'$ frames $\mathbf{I}'_{\text{novel}}$ and poses $\mathbf{V}'_{\text{novel}}$ are selected within novel frames $\mathbf{I}_{\text{novel}}$ and poses $\mathbf{V}_{\text{novel}}$ according to minimum distance to the known poses:
\begin{equation}
      \mathbf{I}'_{\text{novel}}, \mathbf{V}'_{\text{novel}} = \underset{\mathbf{I}_{\text{novel}}', \mathbf{V}_{\text{novel}}'}{\text{argmin}} \sum_{v_{\text{novel}}^{(p)} \in \mathbf{V}_{\text{novel}}}^{P'} \min_{v \in \mathbf{V}_{\text{known, gen}}} \| v_{\text{novel}}^{(p)} - v \|.
\label{eq:iterative_selection}
\end{equation}
Then, we append the selected novel poses and frames into the known set, $\mathbf{I}_{\text{known}} = \mathbf{I}_{\text{known}} \cup \mathbf{I'}_{\text{novel}}, \mathbf{V}_{\text{known}} = \mathbf{V}_{\text{known}} \cup \mathbf{V'}_{\text{novel}}$.
At the end of an iteration, we train the layered GS on these known poses and images to learn the consistent content from the video prior progressively. 
This layered GS optimization is introduced in Sec.~\ref{subsec:uncertainty}.
We empirically define 300 to 400 unknown camera poses in total and repeat this process multiple times until all cameras become known.
Please refer to supp. material for camera definition.

\subsection{Uncertainty-aware Training}
\label{subsec:uncertainty}

To avoid error drifting during mutual conditioning and achieve a robust reconstruction, an uncertainty measurement is required to distinguish the reliable parts of generated novel views.
In our task, uncertainty arises due to limited constraints from sparse-view input, which exhibits excessive degrees of freedom in the content of generated novel views, as shown in Fig.~\ref{fig:method_uncertainty},
\ie, diffusion hallucinates incorrect geometry and appearance in 
high-freedom regions.
Inspired by NeRF Ensembles~\cite{sunderhauf2023density},
we exploit epistemic uncertainty~\cite{lakshminarayanan2017simple} to model this hallucination.
Specifically, we generate multiple images from the same viewpoint, each conditioned on either unperturbed or perturbed point cloud renderings.
As depicted in Fig.~\ref{fig:method_uncertainty}, we measure the variance across these generated images as epistemic uncertainty, which reflects the model's lack of knowledge in under-constrained regions.
The perturbation process is based on the L1 difference between the generated views and their corresponding unperturbed conditions. Perturbations are then selectively applied to areas of conditions where this difference exceeds a predefined threshold $\beta$.

Moreover, the uncertainty map is combined with some loss terms to supervise the training.
We employ the photometric loss $\mathcal{L}_c$ and perceptual loss $\mathcal{L}_{\text{lpips}}$ between rendered images and generated images, and a cosine loss $\mathcal{L}_{n}$ between rendered normal and monocular normal at input viewpoints.
Additionally, we use binary entropy loss $\mathcal{L}_{m}$ between alpha maps of the front GS and front-layer masks to retain the shape.
We also exploit some regularization terms $\mathcal{L}_{\text{reg}}$, such as distortion loss and normal consistency.

\vspace{-0.3em}
\begin{equation}
      \mathcal{L} = \mathbf{u} *[\lambda_c (\mathcal{L}_c + \mathcal{L}_{\text{lpips}}) + \lambda_m \mathcal{L}_m  + \lambda_n \mathcal{L}_n +  \lambda_{\text{reg}} \mathcal{L}_{\text{reg}}],
\label{eq:final_loss}
\end{equation}
where $\lambda_{m}=1.0$, $\lambda_{n}=0.25$. $\lambda_{\text{reg}}$ is 100 for distortion loss and 0.25 for normal consistency loss.
As the inconsistency will accumulate as the camera deviates from the input view mentioned in Sec.~\ref{subsec:iterative}, we relate the $\lambda_{c}$ with the distance $x$ to the input sparse views, $\lambda_{c} = f(x) = \max(0.5*e^{(-20*x)}, 0.05)$.
During training, the densification process is deactivated to prevent densifying Gaussians based on inaccurate positional gradients arising from inconsistencies of generated views.
Please refer to the supp. material for training details.

\ssecspace

\section{Experiments}

\begin{figure*}[!t]
\centering
\vspace{-1.5em}
\includegraphics[width=0.95\linewidth, trim={0 0 0 0}, clip]{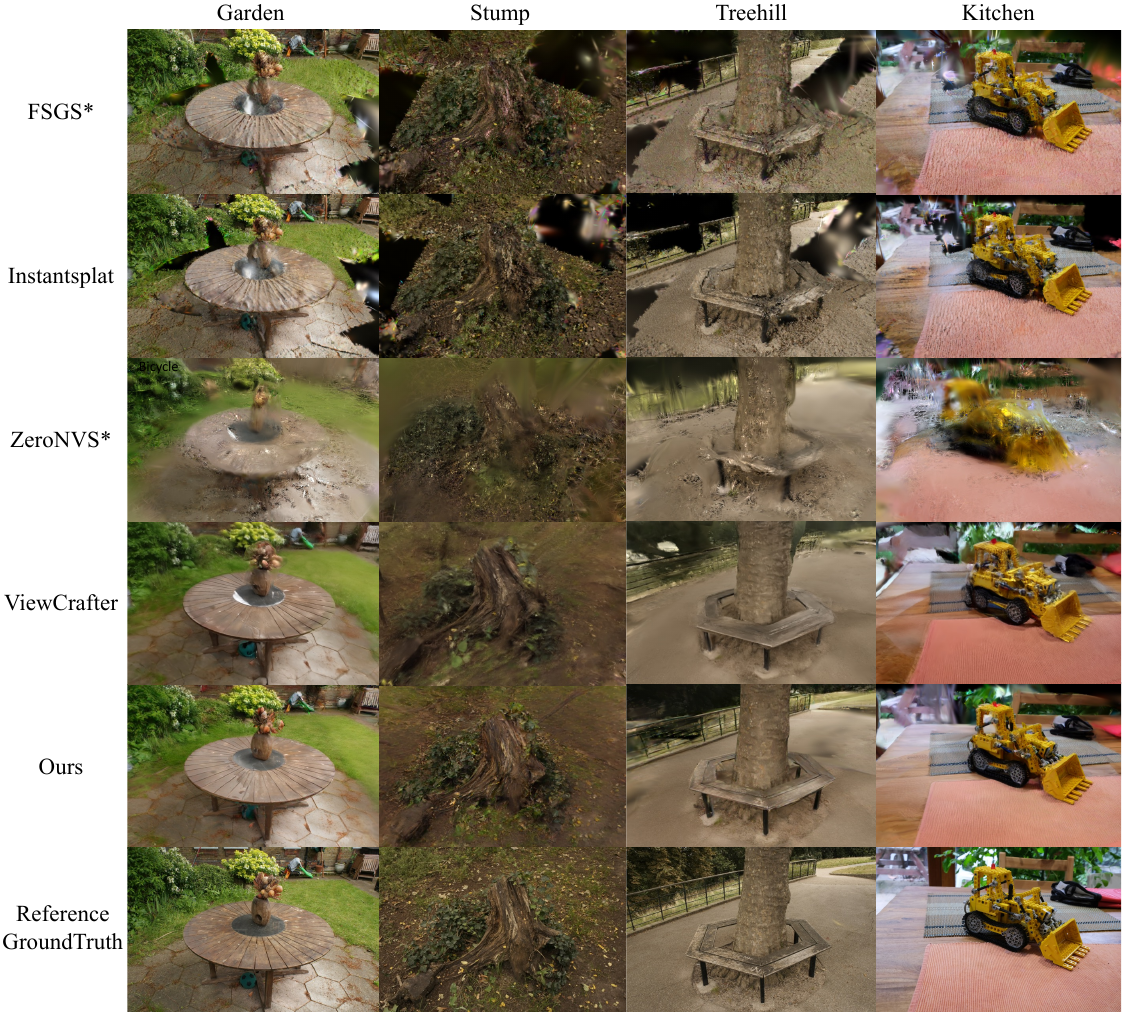}
\vspace{-1.0em}
\caption{\textbf{Comparison on Mip-NeRF 360$^{\circ}$~\cite{barron2022mipnerf360} Dataset on the 3-View Setting.} We qualitatively compare rendering quality with FSGS*~\cite{fsgs}, 
InstantSplat~\cite{fan2024instantsplat}, ZeroNVS*~\cite{sargent2024zeronvs}, ViewCrafter~\cite{yu2024viewcrafter} given 3 input views. 
}
 \vspace{-1.0em}
\label{fig:mipnerf_comp}
\end{figure*}

\begin{figure*}[!t]
\centering
\vspace{-1.5em}
\includegraphics[width=1.0\linewidth, trim={0 0 0 0}, clip]{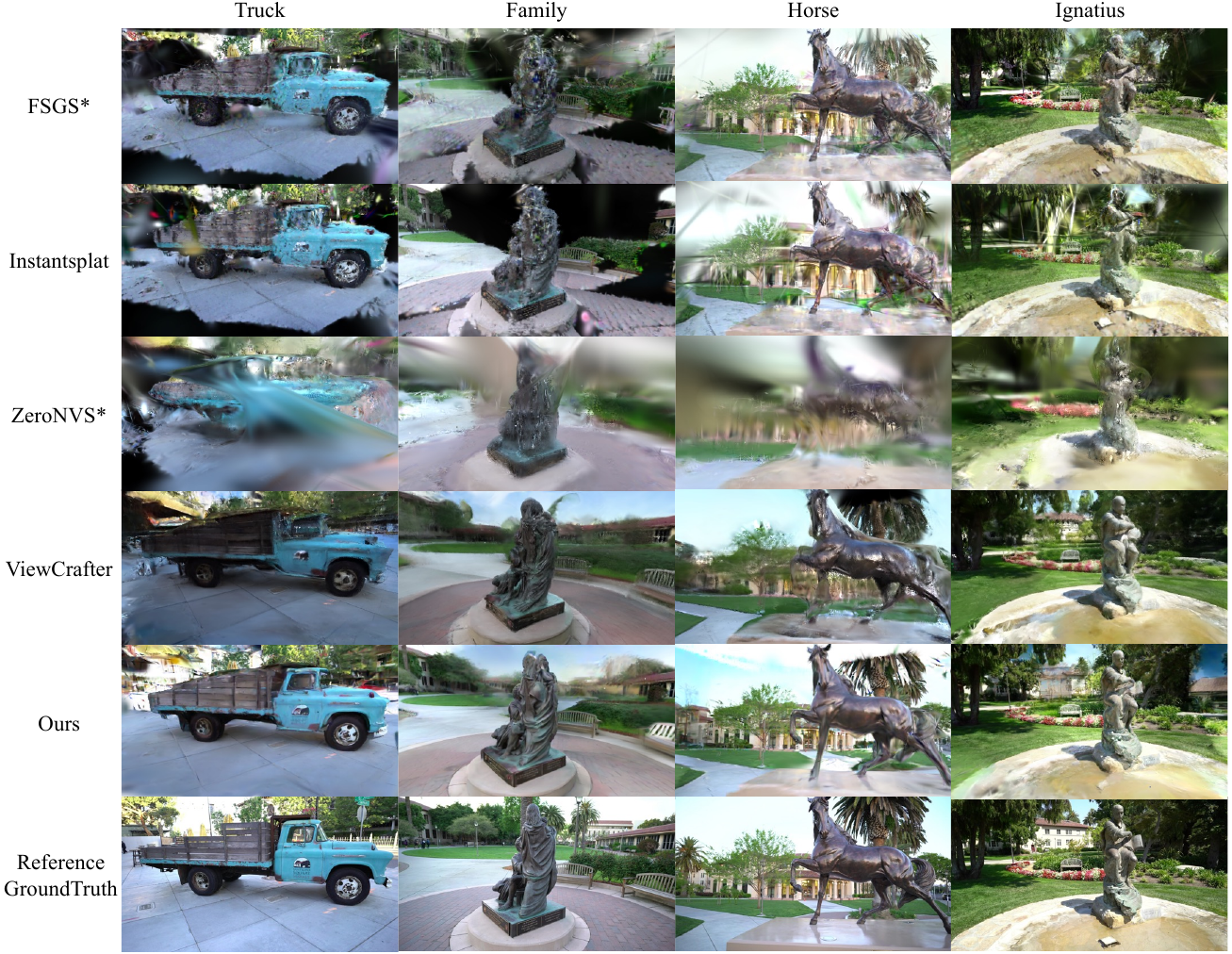}
\vspace{-2.0em}
\caption{\textbf{Comparison on Tanks and Temples~\cite{tanksandtemples} Dataset on the 4-View Setting.} We qualitatively compare rendering quality with FSGS*~\cite{fsgs}, 
InstantSplat~\cite{fan2024instantsplat}, ZeroNVS*~\cite{sargent2024zeronvs}, ViewCrafter~\cite{yu2024viewcrafter} given 4 views. 
    }
\label{fig:tnt_comp}
\end{figure*}

\begin{figure*}[!t]
\centering
\vspace{-1.0em}
\includegraphics[width=1.0\linewidth, trim={0 0 0 0}, clip]{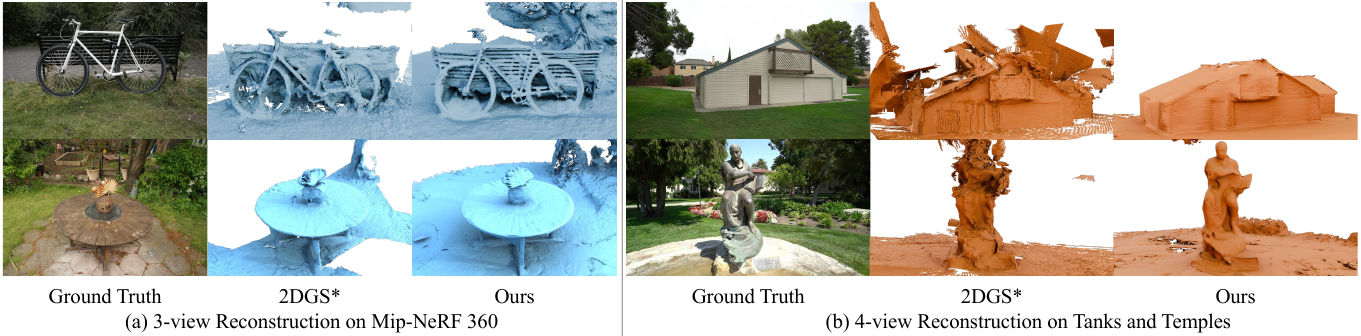}
\vspace{-2.0em}
\caption{\textbf{Comparison on 3D Surface Reconstruction.} We qualitatively compare surface reconstruction with 2DGS~\cite{huang20242d} on (a) Bicycle (top) and Garden (down) from Mip-NeRF 360$^{\circ}$~~\cite{barron2022mipnerf360} dataset and (b) Barn (top) and Ignatius (down) from Tanks and Temples dataset~\cite{tanksandtemples}.
    }
\vspace{-1.0em}
\label{fig:mesh_comp}
\end{figure*}

\subsection{Dataset \& Baselines}

\noindent\textbf{Dataset.} We evaluate our method on two unbounded 360$^{\circ}$ datasets: Mip-NeRF 360~\cite{barron2022mipnerf360} and Tanks and Temples~\cite{tanksandtemples}.
For Mip-NeRF 360~\cite{barron2022mipnerf360}, we choose 8 out of 9 sequences containing indoor and outdoor scenes, excluding the Flower scene as DUSt3R~\cite{dust3r} fails using sparse views.
We follow 3DGS's~\cite{3dgs} train-test split and then select 3, 6, 9 views from its training split as training images, and evaluate novel views on its test split.
For Tanks and Temples~\cite{tanksandtemples}, we use 6 unbounded outdoor scenes, each captured in 360$^{\circ}$.
Similarly, we uniformly split the train and test set by sampling every 8th image as the test set, and the remaining images are training images.
As 3 views are insufficient to capture the 360-degree scenes,
we select 4, 6, and 9 views from the training images as sparse input and evaluate novel views on the test split. We align estimated poses from DUSt3R~\cite{dust3r} with Colmap~\cite{schonberger2016structure} poses recovered using all images for evaluation. Please refer to the supp. material for more explanations.

\noindent\textbf{Baselines.} We compare with the following state-of-the-art methods\cite{fsgs,fan2024instantsplat,sargent2024zeronvs,yu2024viewcrafter}. 
FSGS~\cite{fsgs} is a 3DGS-based method that regularizes the Gaussian optimization with monocular depth prior.
ZeroNVS~\cite{sargent2024zeronvs} trains a diffusion model for the single-view reconstruction in unbounded scenes.
We follow ReconFusion~\cite{wu2024reconfusion} to adapt it to multi-view input.
For fair comparisons, we use the same input point cloud and camera poses from DUSt3R for FSGS* and ZeroNVS*.
InstantSplat~\cite{fan2024instantsplat} initializes Gaussian primitives from DUSt3R's point cloud and optimizes 3D Gaussians and camera poses jointly.
ViewCrafter~\cite{yu2024viewcrafter} fine-tunes a video diffusion model conditioned on point cloud rendering from DUSt3R and trains a final 3DGS~\cite{3dgs} based on generated views and sparse input.
Please refer to the supp. material for more details on baseline training.

\subsection{Novel View Synthesis}

We now evaluate our method on the Mip-NeRF 360 dataset~\cite{barron2022mipnerf360}. As shown in Tab.~\ref{tab:comp_mipnerf}, our method outperforms all baselines across all metrics. As shown in Fig.~\ref{fig:mipnerf_comp}, this dataset is particularly challenging as it contains many fine-grained structures, like the delicate flowers in the Garden. FSGS~\cite{fsgs} and InstantSplat~\cite{fan2024instantsplat} show strong artifacts, such as "foggy" geometry and needle-like distorted Gaussians in the background. Their reconstruction quality is limited by the noisy point cloud from DUSt3R, especially for fine structures. In contrast, our Bootstrap Optimization corrects these errors, restoring details like the flowers in the \textit{Garden} scene and the leaves in the \textit{Stump} scene. ZeroNVS~\cite{sargent2024zeronvs} fails to generate consistent novel views for 3DGS training, while ViewCrafter~\cite{yu2024viewcrafter} has similar issues. Inconsistent novel views from the video diffusion model also impair quality, resulting in blurred renderings and view-dependent artifacts, as seen in the flowers of \textit{Garden} scene and split wheels of Lego in \textit{Kitchen} scene. Our Iterative Fusion effectively mitigates inconsistencies, improving reconstruction.

We further evaluate our method on the Tanks and Temples dataset~\cite{tanksandtemples}, which presents larger scenes and significant depth variations, challenging reconstruction from sparse views. Similar to our observation above, our method outperforms all baselines, as shown in Tab.~\ref{tab:comp_tnt}. 
We observe similar artifacts for baseline methods in Fig.~\ref{fig:tnt_comp}: FSGS~\cite{fsgs} and InstantSplat~\cite{fan2024instantsplat} produce needle-like Gaussian for the foreground object due to noisy point cloud initialization. Further, they failed to complete the background regions as they did not employ generative priors. ZeroNVS\cite{sargent2024zeronvs} produces overly blurred images, while ViewCrafter~\cite{yu2024viewcrafter} produces distorted geometry due to noise in the point cloud initialization and inconsistency in the generative models. For instance, the leg of the \textit{horse} is missing in ViewCrafter's rendering, and the sky sticks with the foreground objects. By contrast, our multi-layer representation mitigates depth ambiguity, preserving foreground integrity. Our bootstrap optimization improves the scene's geometry, resulting in better NVS results.

\begin{table}[t!]
\vspace{-0.5em}
\centering
\resizebox{0.80\linewidth}{!}{
\tabcolsep 10pt
\begin{tabular}{lccc}
\hline
\textbf{Methods} & \textbf{4 views} & \textbf{6 views} & \textbf{9 views} \\
\hline
2DGS~\cite{huang20242d} & 0.284 & 0.367 & 0.418 \\
Ours  & \textbf{0.423} & \textbf{0.430} & \textbf{0.439} \\
\hline
\end{tabular}
}
\vspace{-0.7em}
\caption{\textbf{Comparison on 3D Surface Reconstruction.} We compare with 2DGS~\cite{huang20242d} on four scenes of Tanks and Temples~\cite{tanksandtemples}. 
The evaluation metric is F1-score and higher is better.
}
\vspace{-0.5em}
\label{tab:comp_mesh_tnt}
\end{table}

\begin{table}[t!]
\centering
\resizebox{1.0\linewidth}{!}{
\tabcolsep 10pt
\begin{tabular}{lccc}
\hline
\textbf{Settings} & \textbf{PSNR}$\uparrow$ & \textbf{SSIM}$\uparrow$ & \textbf{LPIPS}$\downarrow$ \\
\hline
Baseline & 16.16 & 0.289 & 0.532 \\
+ Multi-layer (Sec.~\ref{subsec:multi_3dgs}) & 16.56 & 0.308 & 0.501 \\
+ Bootstrap Opt. (Sec.~\ref{subsec:bootstrap}) & 16.65 & 0.314 & 0.491 \\
+ Iterative Fusion (Sec.~\ref{subsec:iterative}) & 16.81 & 0.317 & 0.484 \\
+ Uncertainty (Ours, Sec.~\ref{subsec:uncertainty}) & \textbf{16.95} & \textbf{0.321} & \textbf{0.480} \\
\hline
\end{tabular}
}
\vspace{-0.5em}
\caption{\textbf{Ablation on Design Choices.} We perform ablation studies on the baseline with different designs of our framework.}
\label{tab:ablation}
\vspace{-1.0em}
\end{table}

\ssecspace
\subsection{Geometry Evaluation}

We also evaluate surface reconstruction accuracy against 2DGS~\cite{huang20242d}.
We use camera poses and point clouds from DUSt3R~\cite{dust3r,mast3r} as input for 2DGS, denoted as 2DGS*.
For quantitative comparisons, we conduct the experiment on four scenes with ground-truth mesh in Tanks and Temples~\cite{tanksandtemples}, including Barn, Ignatius, Caterpillar, and Truck.
As sparse-view reconstruction is error-prone, we enlarge the error threshold 10 times for all methods to compute the F1-score.
As is shown in Tab.~\ref{tab:comp_mesh_tnt}, our method achieves superior results to 2DGS~\cite{huang20242d}.
We show qualitative comparisons on Mip-NeRF 360~\cite{barron2022mipnerf360} and Tanks and Temples~\cite{tanksandtemples} in Fig.~\ref{fig:mesh_comp}. 
2DGS~\cite{huang20242d} can not reconstruct the geometry of unseen regions; as a consequence, there are large holes on the ground and the background as shown in Fig.~\ref{fig:mesh_comp}(a). Moreover, noisy point clouds from dense stereo models~\cite{dust3r,mast3r} lead to error-prone densification, resulting in artifacts in the geometry, particularly in the sky region, where 2DGS~\cite{huang20242d} grows Gaussian points on the foreground to minimize appearance loss, see \textit{Barn} scene in Fig.~\ref{fig:mesh_comp}(b). By contrast, through Bootstrap Optimization and Iterative Fusion, we obtain a clean and detailed point cloud for initialization, allowing smoother distribution of Gaussian points with less noise at the seen regions and complete geometry at the unseen regions.

\ssecspace
\subsection{Ablation Study}
\begin{figure}[!t]
\centering
\vspace{-1.0em}
\includegraphics[width=1.0\linewidth, trim={0 0 0 0}, clip]{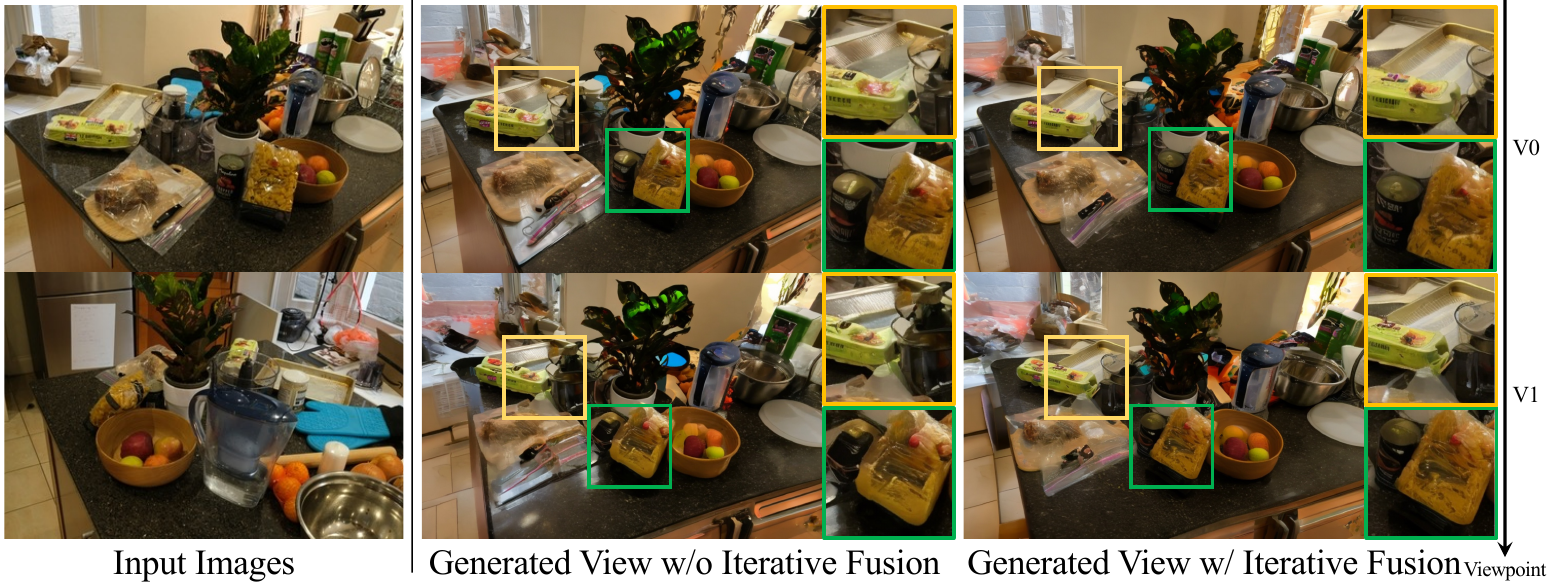}
\vspace{-2.0em}
\caption{
\textbf{Ablation on Iterative Fusion.}
We compare the rendering quality of different viewpoints when ablating the iterative fusion (Sec.~\ref{subsec:iterative}) on the Counter of Mip-NeRF 360$^{\circ}$~\cite{barron2022mipnerf360} dataset.
    }
\label{fig:ablation}
\vspace{-1.0em}
\end{figure}
We conduct ablations using \textit{Bicycle} and \textit{Garden} from Mip-NeRF 360~\cite{barron2022mipnerf360} with 3 input views.
Given stereo-reconstructed point cloud and poses, ``Baseline'' firstly generates images under the same unknown poses as ours, then optimizes a final 2DGS~\cite{huang20242d} on generated and input images.
Adding multiple layers into GS optimization boosts the rendering quality.
Bootstrap Optimization provides a cleaner and more detailed geometry for Gaussian Splatting to capture detailed geometry and appearance. Iterative Fusion and Uncertainty-aware Training generate more consistent novel views, leading to better results at inconsistent regions, which further improves the rendering quality. As shown in Tab.~\ref{tab:ablation}, each component can effectively 
enhance the quality of reconstruction. 
We also compare generated novel views from the video diffusion model on 3 views of Counter from Mip-NeRF 360~\cite{barron2022mipnerf360}. 
As shown in Fig.~\ref{fig:ablation}, the generated results without Iteration Fusion suffer inconsistency across multiple viewpoints, 
while with Iteration Fusion, video diffusion model can generate consistent and high-quality novel views.

\ssecspace

\section{Conclusion}
We have presented a novel framework for unposed and extremely sparse-view 3D reconstruction in unbounded 360 scenes.
We build a layered GS upon the reconstruction of dense stereo reconstruction model~\cite{dust3r,mast3r} and use bootstrap optimization to generate a clean and detailed geometry from sparse views.
Besides, we propose an iterative fusion of reconstruction and generation~\cite{yu2024viewcrafter} to enable mutual conditioning and enhancement, and an uncertainty-aware training for robust reconstruction.
It is worth noting that Free360 is a general framework that is both complementary and orthogonal to video diffusion models. A future direction is incorporating an enhanced video diffusion model into our framework.
As a limitation, we cannot solve scenes with extensive repetitive texture where stereo reconstruction model~\cite{dust3r,mast3r} fails for faithful global reconstructions.

\noindent\textbf{Acknowledgment:} This work was partially supported by the NSFC (No.~62441222), Information Technology Center and State Key Lab of CAD\&CG, Zhejiang University.

{
    \small
    \bibliographystyle{ieeenat_fullname}
    \bibliography{main}
}

\clearpage

\appendix

\renewcommand\thesection{\Alph{section}}
\renewcommand\thetable{\Alph{table}}
\renewcommand\thefigure{\Alph{figure}}

\begin{strip}
\begin{center}
{\huge \bf Supplementary Material}
\end{center}
\end{strip}

\maketitle

In this supplementary material, we first present detailed implementation aspects in Section~\ref{sec:impl}.
More experimental details are shown in Section~\ref{sec:exp_detail}. 
We show more comparisons in the Sec.~\ref{sec:more_exp}.
Additionally, we include a short video summarizing the method with video results, and an offline webpage for interactive visualization of our whole results and comparisons.

\section{Implementation Details}
\label{sec:impl}
For the fine-grained front and back layer masks, we annotate the maximum depth of the front layer in monocular depth~\cite{depthanythingv2} of each view.
The pixels are selected into the front layer if their depth is smaller than the annotated maximum depth.
We build Free360 upon the 2DGS~\cite{huang20242d} framework.
We follow the version implemented in the StableNormal~\cite{ye2024stablenormal}.
We use default settings in dense stereo reconstruction models~\cite{dust3r, mast3r} and use the filtered point cloud by predicted confidence map.
We transform the world origin to the center of the scene, which is determined by the center depth of the first image.
Besides, we rescale the cameras to fit within a sphere of radius 2.

In reconstruction bootstrap optimization, we downsample the point cloud of the front layer before initializing its Gaussian primitives.
We initialize the front layer's Gaussian primitives using its point cloud and train for 10,000 iterations based on the loss defined in Eq.~(\textcolor{red}{1}).
We enable densification from the 166-th iteration to 5000-th iteration.

In the iterative fusion of reconstruction and generation, we define the unknown cameras in two ways. 
First, we interpolate the poses between input sparse views in the cubic spline interpolator.
Second way is to define a target camera pose by jittering the position of an input camera pose while orienting its rotation to face the world origin, and interpolate the poses between the target pose and closet input pose.
We empirically define 300 to 400 unknown camera poses in total from these two ways.
We use ViewCrafter~\cite{yu2024viewcrafter} to generate the 25 frames each time with the resolution 1024$\times$576. 

In uncertainty-aware training, we set the maximum L1 difference $\beta$ between conditions and generations as 0.2.
We train the Gaussian primitives of the front layer and back layer using Eq.~(\textcolor{red}{3}) in 10000 iterations without densification.
All experiments are conducted on an NVIDIA RTX 6000 GPU.

 \begin{figure*}[!t]
\centering
\includegraphics[width=0.95\linewidth, trim={0 0 0 0}, clip]{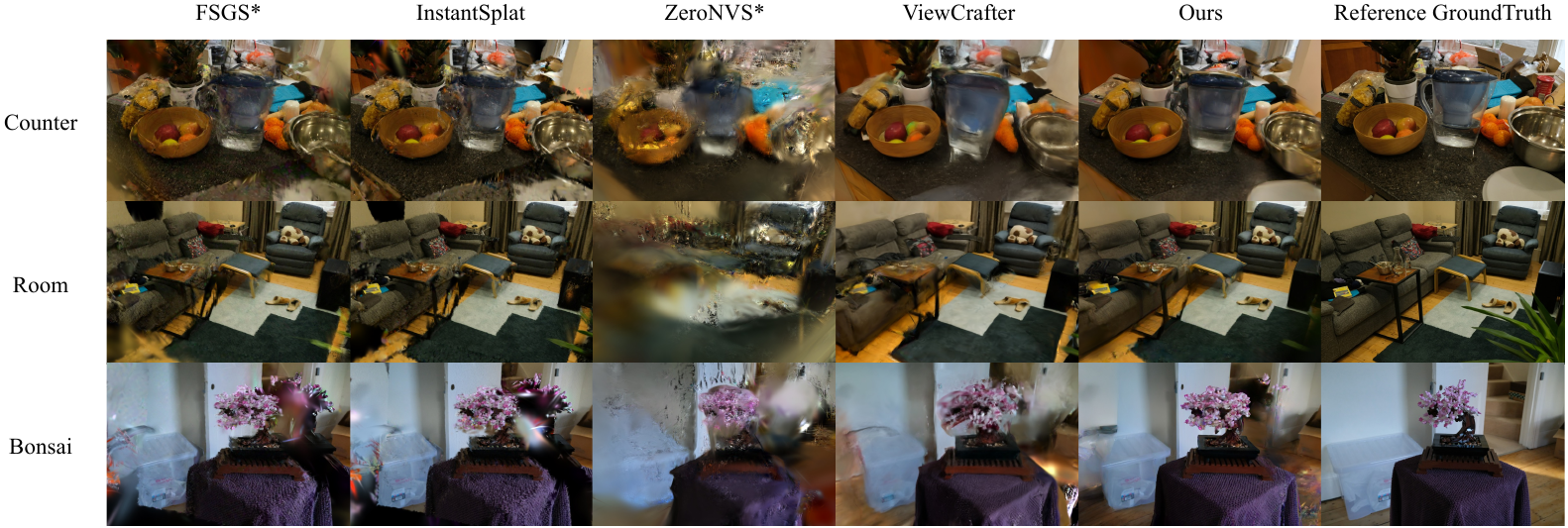}
\caption{\textbf{Comparison on Mip-NeRF 360$^{\circ}$~\cite{barron2022mipnerf360} Dataset on the 3-View Setting.} We qualitatively compare rendering quality with FSGS*~\cite{fsgs}, 
InstantSplat~\cite{fan2024instantsplat}, ZeroNVS*~\cite{sargent2024zeronvs}, ViewCrafter~\cite{yu2024viewcrafter} given 3 input views. 
}
\label{fig:mipnerf_comp_supp}
\end{figure*}

\begin{figure*}[!t]
\centering
\includegraphics[width=1.0\linewidth, trim={0 0 0 0}, clip]{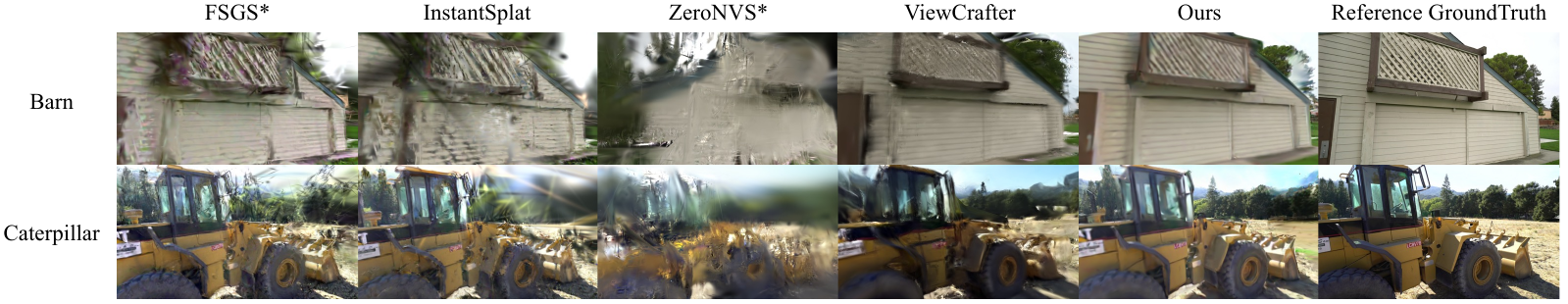}
\caption{\textbf{Comparison on Tanks and Temples~\cite{tanksandtemples} Dataset on the 4-View Setting.} We qualitatively compare rendering quality with FSGS*~\cite{fsgs}, 
InstantSplat~\cite{fan2024instantsplat}, ZeroNVS*~\cite{sargent2024zeronvs}, ViewCrafter~\cite{yu2024viewcrafter} given 4 views. 
    }
\label{fig:tnt_comp_supp}
\end{figure*}

\begin{figure*}[!t]
\centering
\includegraphics[width=1.0\linewidth, trim={0 0 0 0}, clip]{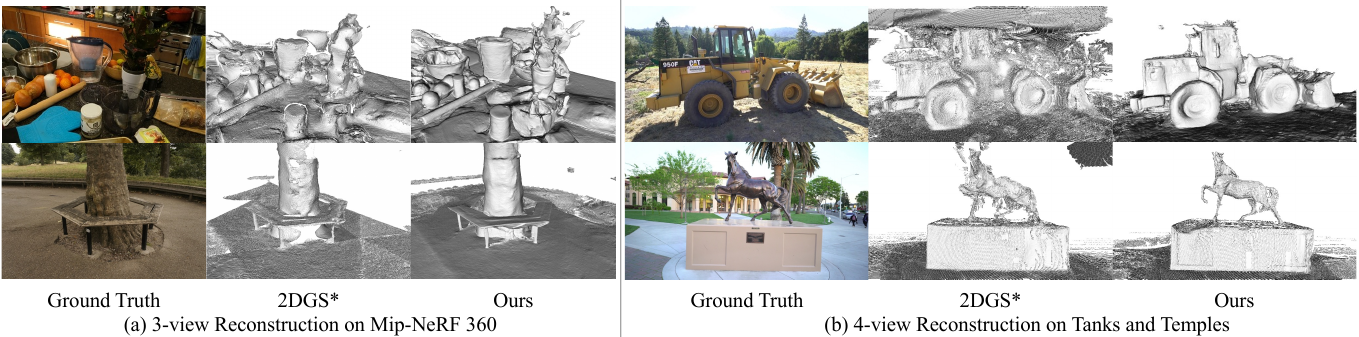}
\caption{\textbf{Comparison on 3D Surface Reconstruction.} We qualitatively compare surface reconstruction with 2DGS~\cite{huang20242d} on (a) Counter (top) and Treehill (down) from Mip-NeRF 360$^{\circ}$~~\cite{barron2022mipnerf360} dataset and (b) Caterpillar (top) and Horse (down) from Tanks and Temples dataset~\cite{tanksandtemples}.
    }
\label{fig:mesh_comp_supp}
\end{figure*}

\section{Experimental Details}
\label{sec:exp_detail}
\subsection{Dataset}
We double the official downsample factor used by 3DGS~\cite{3dgs} in Mip-NerF 360~\cite{barron2022mipnerf360}.
For Tanks and Temples~\cite{tanksandtemples}, we use the processed data from PixelGS~\cite{zhang2024pixelgs} and downsample the images by a factor of 2.
To evaluate the metrics between rendered images and ground-truth images, we follow the InstantSplat~\cite{fan2024instantsplat} to align the estimated poses from stereo reconstruction model~\cite{dust3r, mast3r} to the ground-truth poses.
Initially, a coarse alignment is obtained through rigid point registration between the estimated and ground-truth camera positions at the training viewpoints.
Subsequently, for each rendered image, we fix the Gaussian primitives, and a test-time optimization is performed on the camera pose by minimizing the L1 difference between the rendered and ground-truth images. 
This optimization is executed for 500 iterations using the Adam optimizer~\cite{kingma2014adam}, with a learning rate of 0.0003 for position and 0.0001 for rotation.

\subsection{Baselines}
All compared methods use the same camera poses and point clouds from dense stereo reconstruction~\cite{dust3r,mast3r}.
Since ZeroNVS~\cite{sargent2024zeronvs} and ViewCrafter~\cite{yu2024viewcrafter} are agnostic to the reconstruction backbone, we adopt the same 2DGS backbone~\cite{huang20242d,ye2024stablenormal} for both methods to ensure a fair and rigorous comparison.
In geometry evaluation, the same 2DGS~\cite{huang20242d,ye2024stablenormal} backbone is used as the baseline.
For low-overlapped sparse views of the unbounded scene, ViewCrafter~\cite{yu2024viewcrafter} needs to iteratively generate novel views within a small region, utilize Dust3R on generated views to recover the scene's point cloud, and subsequently repeat to generate the next portion of the scene conditioned previously generated image and Dust3R~\cite{dust3r} point cloud.
However,  Dust3R's feed-forward nature struggles with inconsistencies in generated images, leading to inaccurate depths that degrade subsequent generations.
In contrast, our iterative fusion framework integrates an uncertainty-aware GS optimization after each iteration to refine the generative error promptly.
The optimized 3D-consistent GS rendering is used to condition subsequent generations for consistent multi-view generation guiding the next GS optimization.
ViewCrafter~\cite{yu2024viewcrafter} and ZeroNVS~\cite{sargent2024zeronvs} use the same group of unknown cameras to generate novel views as our method.

\section{More Experiments}
\label{sec:more_exp}
\subsection{Novel View Synthesis}

We show more rendering comparisons on the Mip-NeRF360~\cite{barron2022mipnerf360} and Tanks and Temples~\cite{tanksandtemples}, as illustrated in Fig.~\ref{fig:mipnerf_comp_supp} and Fig.~\ref{fig:tnt_comp_supp} respectively.
FSGS~\cite{fsgs} and InstantSplat~\cite{fan2024instantsplat} exhibit severe distortion and needle-like Gaussian artifacts in the rendering results.
ZeroNVS~\cite{sargent2024zeronvs} fails in synthesizing clear novel views due to limited consistency from generative prior.
ViewCrafter~\cite{yu2024viewcrafter} cannot present a detailed and consistent rendering of the scene.
Instead, our method shows the crisp rendering and complete structure of the scene.
\subsection{Geometry Evaluation}
\begin{figure*}[t!]
    \centering
\resizebox{0.90\linewidth}{!}{
    \begin{tabular}{cc}
    2DGS* & Ours\\
        \begin{subfigure}[t]{0.45\textwidth}
            \centering
            \includegraphics[width=\textwidth]{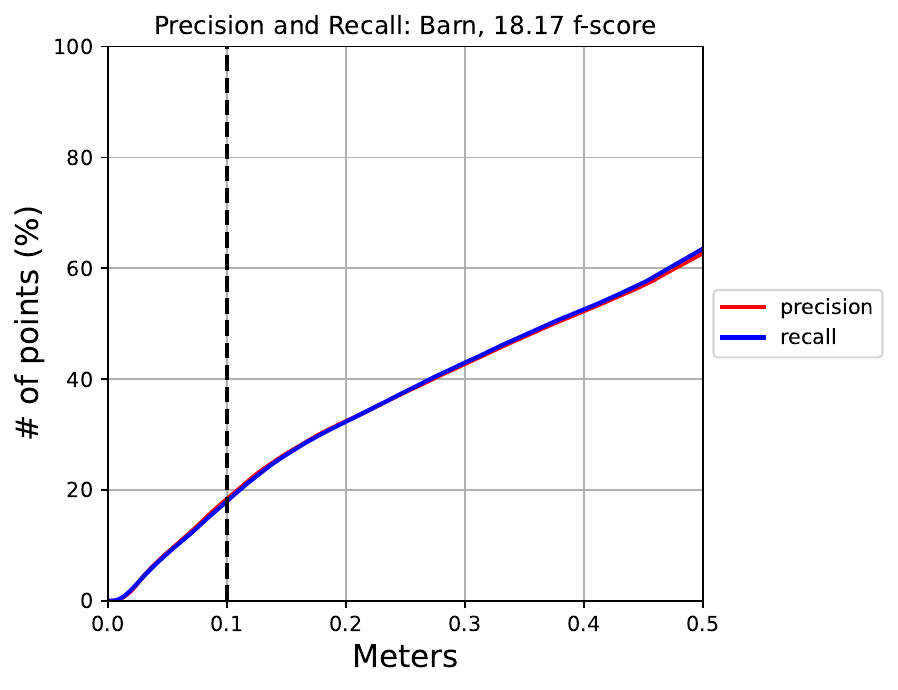}
        \end{subfigure} &
        \begin{subfigure}[t]{0.45\textwidth}
            \centering
            \includegraphics[width=\textwidth]{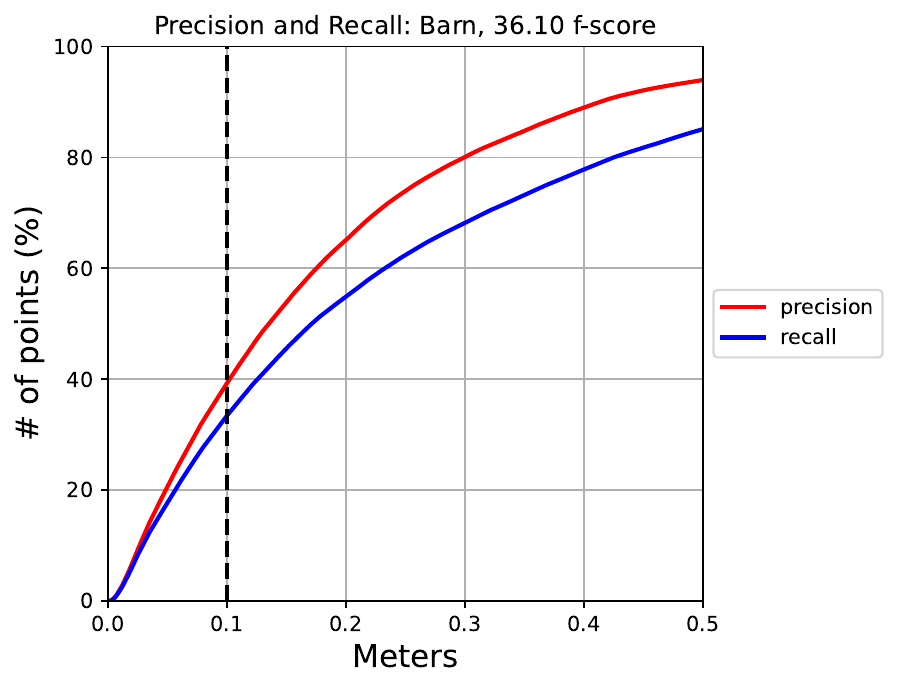}
        \end{subfigure} \\
        
        \begin{subfigure}[t]{0.45\textwidth}
            \centering
            \includegraphics[width=\textwidth]{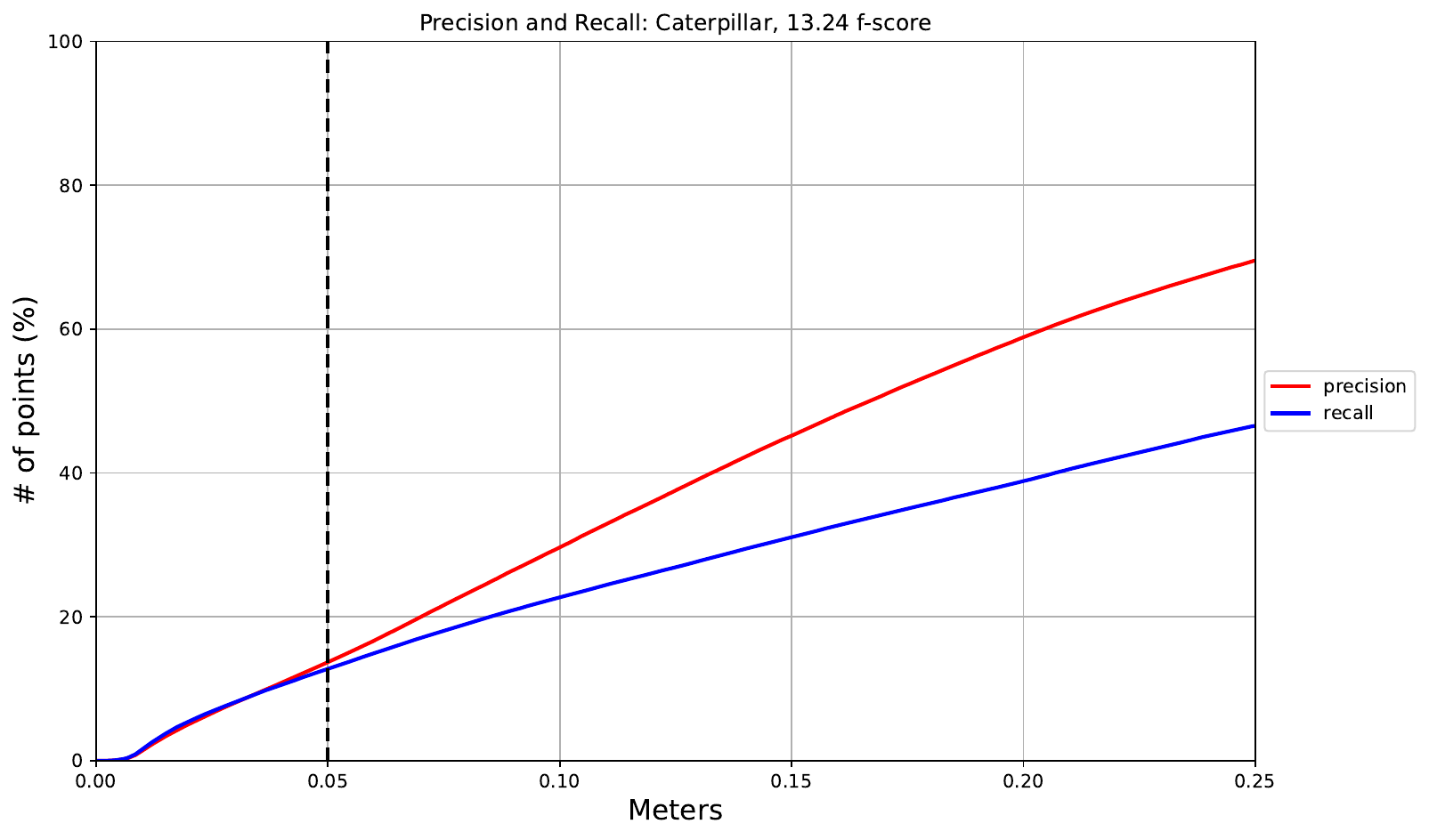}
        \end{subfigure} &
        \begin{subfigure}[t]{0.45\textwidth}
            \centering
            \includegraphics[width=\textwidth]{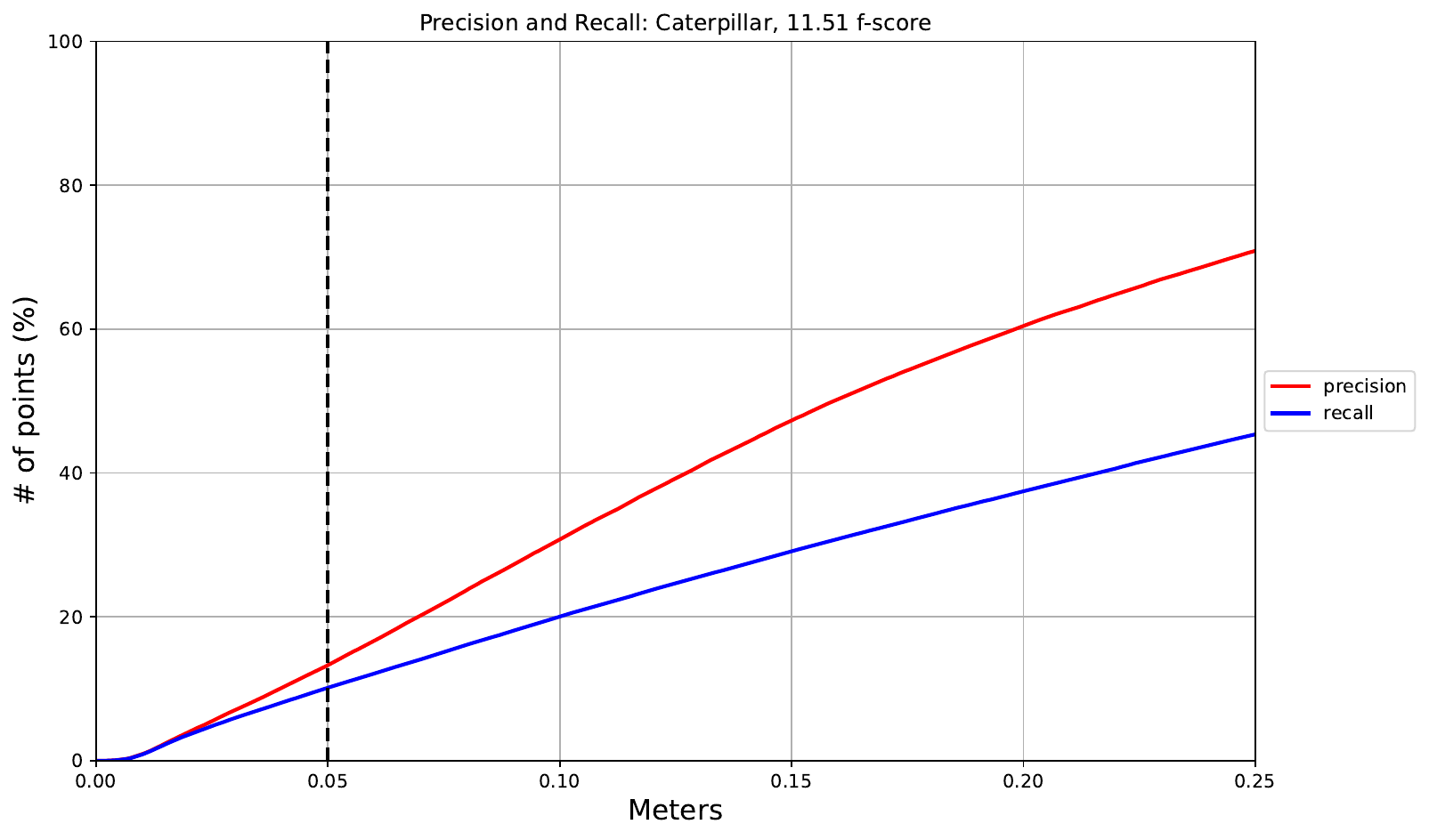}
        \end{subfigure} \\
        
        \begin{subfigure}[t]{0.45\textwidth}
            \centering
            \includegraphics[width=\textwidth]{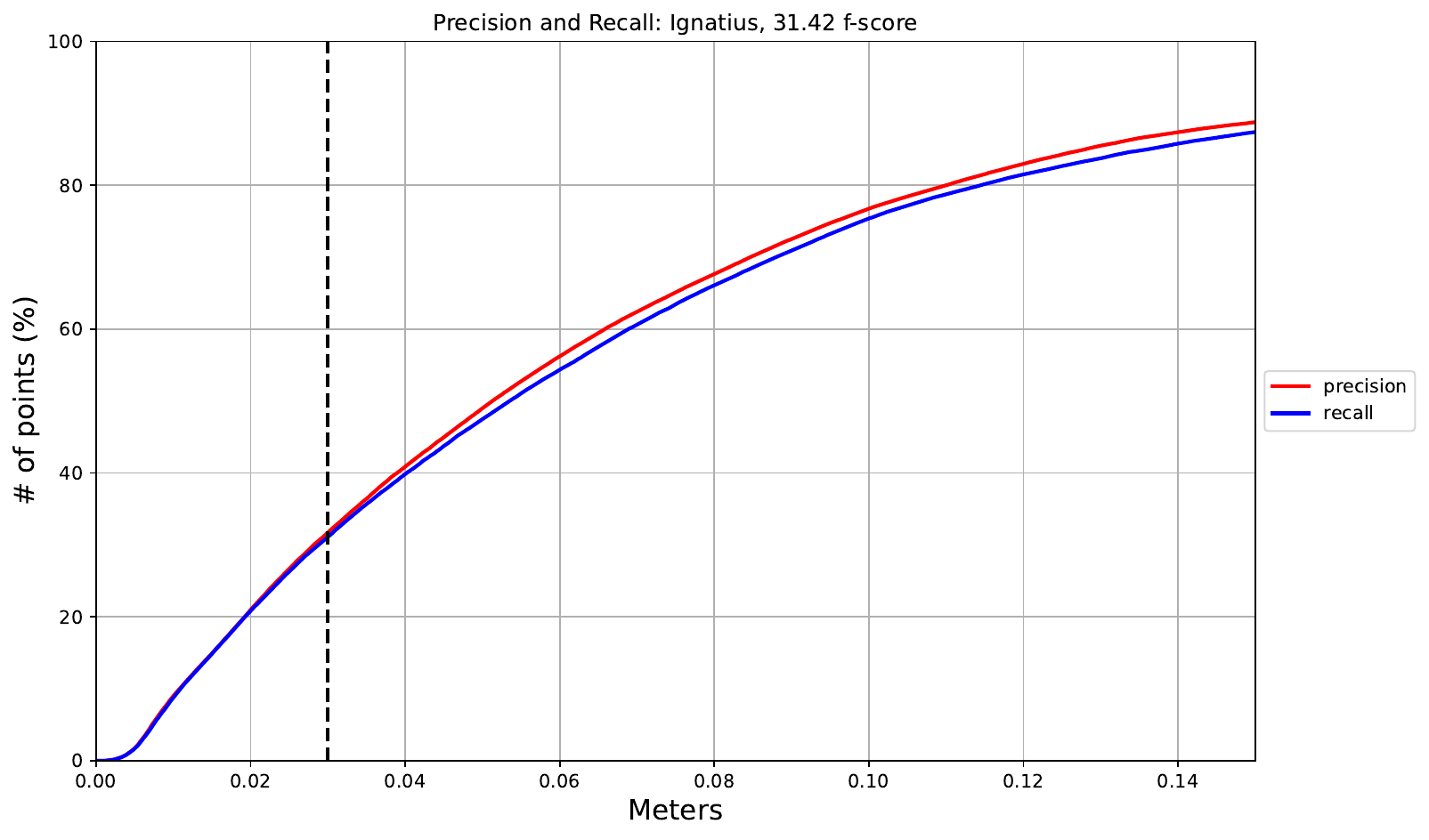}
        \end{subfigure} &
        \begin{subfigure}[t]{0.45\textwidth}
            \centering
            \includegraphics[width=\textwidth]{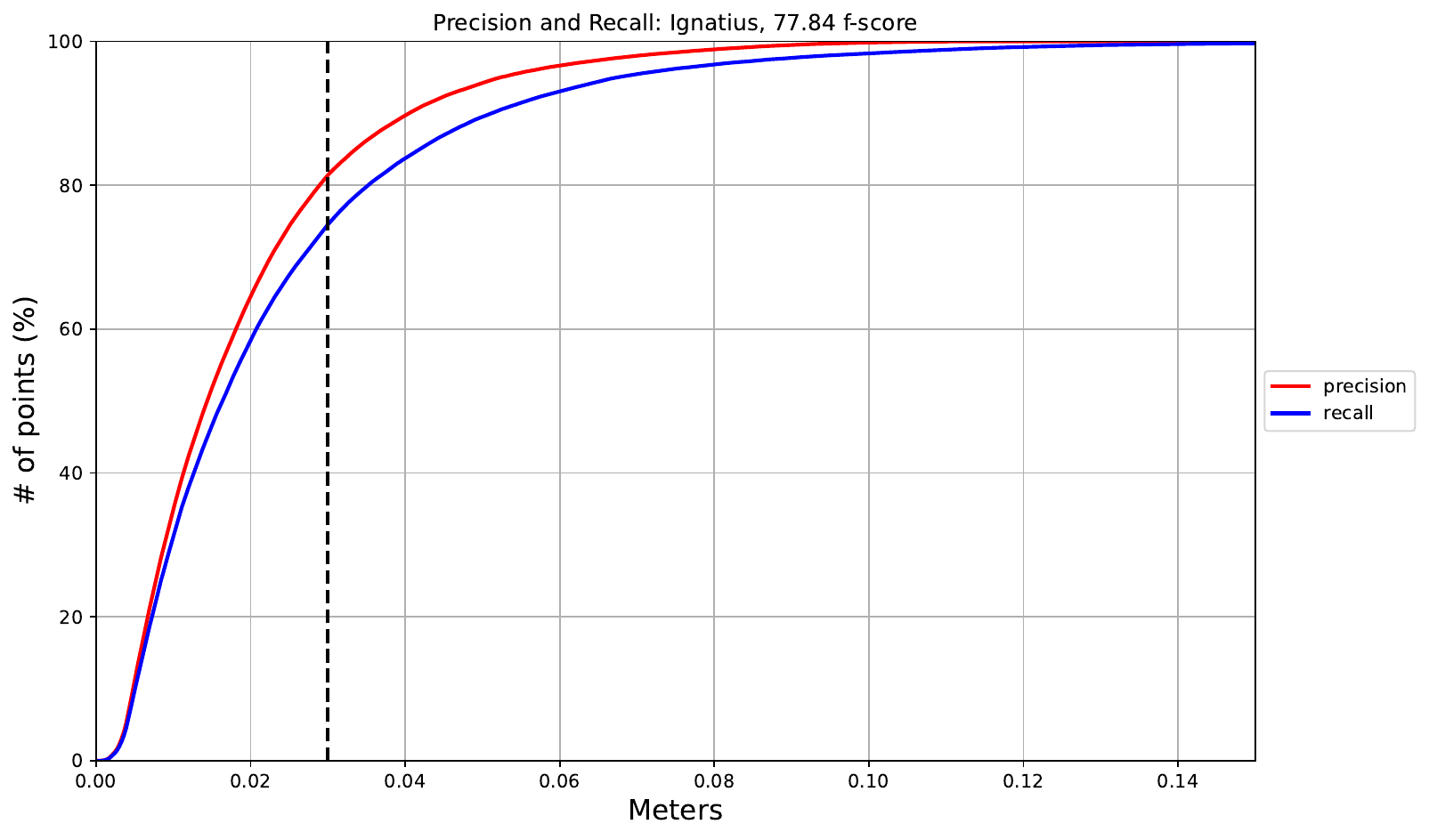}
        \end{subfigure} \\
        
        \begin{subfigure}[t]{0.45\textwidth}
            \centering
            \includegraphics[width=\textwidth]{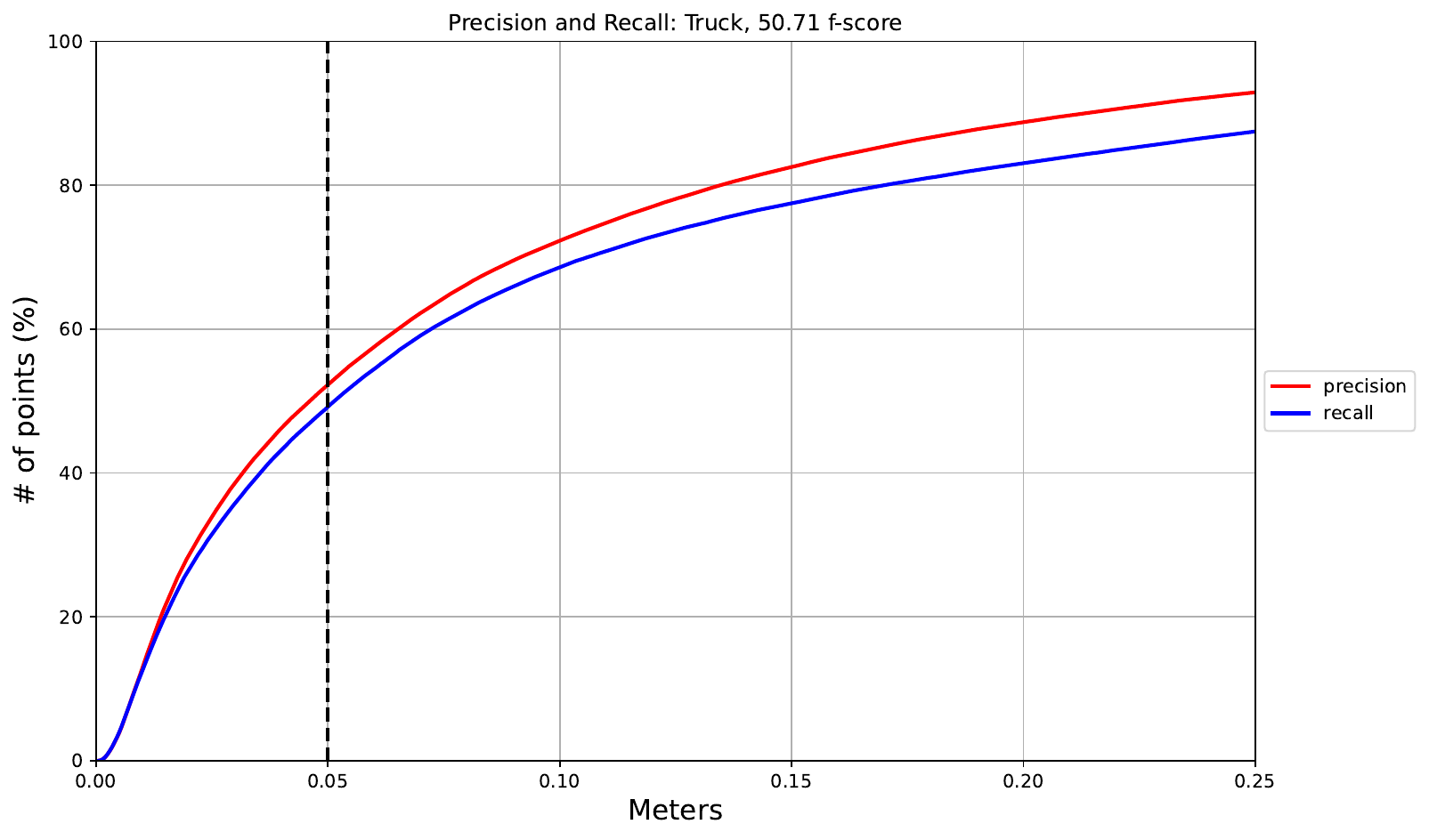}
        \end{subfigure} &
        \begin{subfigure}[t]{0.45\textwidth}
            \centering
            \includegraphics[width=\textwidth]{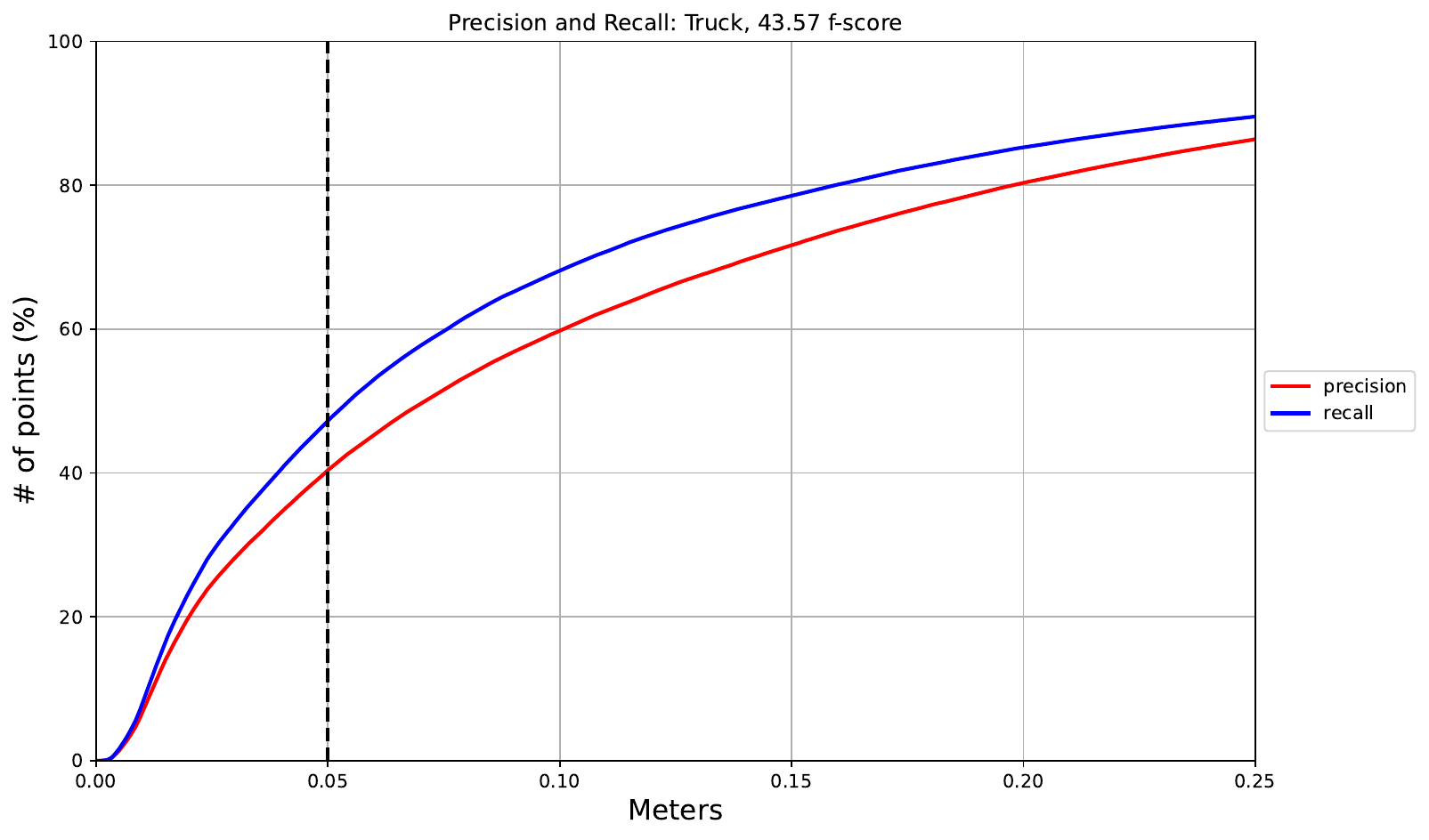}
        \end{subfigure} \\
    \end{tabular}
}
    \caption{\textbf{The Precision and Recall Curves of F1-score Comparisons}. We show detailed precision and recall curves of F1-score comparisons between 2DGS and our method on Barn, Caterpillar, Ignatius, and Truck, given 4 views. The black dotted line in each subfigure denotes the error threshold.}
    \label{fig:f1_curve_comp}
\end{figure*}

We show more geometry evaluation on the Mip-NeRF360~\cite{barron2022mipnerf360} and Tanks and Temples~\cite{tanksandtemples}, as illustrated in Fig.~\ref{fig:mesh_comp_supp}.
Given the sparse views, the geometry from 2DGS has many missing areas and distorted surfaces, such as the holes in Treehill, the missing legs in Horse, and the floater at the top of Caterpillar.
In contrast, our method not only produces a complete and smooth geometry of the scene but also detailed structures.
We show the F1-score precision and recall curves of 2DGS and our method in Fig.~\ref{fig:f1_curve_comp}.
Since the extremely sparse-view surface reconstruction is ambiguous and error-prone, few points lie within the official error threshold in Tanks and Temples~\cite{tanksandtemples}.
To facilitate a clearer comparison, we increase the error threshold by a factor of 10 (represented by the black dotted line in Fig.~\ref{fig:f1_curve_comp}).

\subsection{Ablation on geometry prior}
As shown in Tab.~\ref{table:geo_prior}, we show ablation of geometry priors $\mathcal{L}_d$ and $\mathcal{L}_n$ in Eq.~\textcolor{red}{1}, \textcolor{red}{2} on \textit{Bicycle} and \textit{Garden} (3 views).
The geometry prior improves the rendering quality but is not the key to sparse-view reconstruction.

\begin{table}[t]
\centering
\resizebox{0.7\linewidth}{!}{
\tabcolsep 10pt
\begin{tabular}{lccc}
\hline
\textbf{Settings} & \textbf{PSNR}$\uparrow$ & \textbf{SSIM}$\uparrow$ & \textbf{LPIPS}$\downarrow$ \\
\hline
w/o $\mathcal{L}_d$ \& $\mathcal{L}_n$  & 16.723 & 0.310 & 0.505 \\
w/o $\mathcal{L}_d$ & 16.799 & 0.314 & 0.499 \\
w/o $\mathcal{L}_n$ & 16.825 & 0.313 & 0.498 \\
Ours & \textbf{16.95} & \textbf{0.321} & \textbf{0.480} \\
\hline
\end{tabular}
}
\caption{We perform ablation studies
on the geometric priors in our method.}
\label{table:geo_prior}
\end{table}

\end{document}